\documentclass[10pt,twocolumn,letterpaper]{article}

\usepackage{cvpr}              % To produce the CAMERA-READY version

\usepackage{multirow}
\usepackage[table]{xcolor}
\usepackage{tabularx}
\usepackage{amsthm}
\usepackage[accsupp]{axessibility} 
\definecolor{cvprblue}{rgb}{0.21,0.49,0.74}
\usepackage[pagebackref,breaklinks,colorlinks,allcolors=cvprblue]{hyperref}

\newbool{makePurple}
\setbool{makePurple}{false} 
\newcommand{\conditionalpurple}[1]{%
  \ifbool{makePurple}{%
    \textcolor{purple}{#1}%
  }{%
    #1%
  }%
}
\usepackage[most]{tcolorbox}
\usepackage{xcolor}

\definecolor{grayblue}{RGB}{220,228,238}
\definecolor{lightbluebox}{RGB}{225,238,250}

\def\paperID{*****} % *** Enter the Paper ID here
\def\confName{CVPR}
\def\confYear{2026}

\title{THE MORE, THE MERRIER: CONTRASTIVE FUSION FOR HIGHER-ORDER MULTIMODAL ALIGNMENT}

\author{
Stefanos Koutoupis$^{1,2,}$\thanks{Corresponding author: skoutoupis@ics.forth.gr},
Michaela Areti Zervou$^{1,2}$,
Konstantinos Kontras$^{3}$, \\
Maarten De Vos$^{3}$,
Panagiotis Tsakalides$^{1,2}$,
Grigorios Tsagkatakis$^{1,2}$\\[4pt]
$^{1}$Foundation for Research and Technology-Hellas,\\ 
$^{2}$University of Crete, 
$^{3}$KU Leuven
}

\begin{document}
% \twocolumn[{%
% \renewcommand\twocolumn[1][]{#1}%
\maketitle
% \begin{center}
%     \centering
%     \captionsetup{type=figure}
%     \includegraphics[width=.6\textwidth]{figs/concept_image.png}
%     \captionof{figure}{Test caption}
% \end{center}%
% }]

\begin{abstract}
Learning joint representations across multiple modalities remains a central challenge in multimodal machine learning. Prevailing approaches predominantly operate in pairwise settings, aligning two modalities at a time. While some recent methods aim to capture higher-order interactions among multiple modalities, they often overlook or insufficiently preserve pairwise relationships, limiting their effectiveness on single-modality tasks. In this work, we introduce Contrastive Fusion (ConFu), a framework that jointly embeds both individual modalities and their fused combinations into a unified representation space, where modalities and their fused counterparts are aligned. ConFu extends traditional pairwise contrastive objectives with an additional fused-modality contrastive term, encouraging the joint embedding of modality pairs with a third modality. This formulation enables ConFu to capture higher-order dependencies, such as XOR-like relationships, that cannot be recovered through pairwise alignment alone, while still maintaining strong pairwise correspondence. We evaluate ConFu on synthetic and real-world multimodal benchmarks, assessing its ability to exploit cross-modal complementarity, capture higher-order dependencies, and scale with increasing multimodal complexity. Across these settings, ConFu demonstrates competitive performance on retrieval and classification tasks, while supporting unified one-to-one and two-to-one retrieval within a single contrastive framework. We release our code and dataset at \url{https://github.com/estafons/confu}.
\end{abstract}

\section{Introduction}
\label{sec:intro}
Learning joint representations across multiple modalities is a fundamental goal in multimodal machine learning~\cite{yuan2025survey}. Contrastive frameworks such as CLIP~\cite{radford2021learning} and ALIGN~\cite{jia2021scaling} have demonstrated that aligning paired modalities, such as images and text, enables powerful zero-shot transfer and retrieval. However, these approaches remain intrinsically pairwise, capturing only correlations between two modalities while overlooking the higher-order dependencies that emerge when three or more modalities interact~\cite{guzhov2022audioclip, jia2021scaling, girdhar2023imagebind, zhu2023languagebind, wang2024omnibind, wang2024freebind, zhou2023uni3d, xue2023ulip, xue2024ulip, wang2023connecting, zhang2024extending}. 
% Beyond this structural limitation, multimodal models often exhibit modality competition~\cite{huang2022modality}, where one modality dominates prediction and suppresses others, particularly when signal strength or reliability differs across modalities. For instance, strong visual cues may overshadow weaker auditory ones, leading the joint representation to rely disproportionately on a single source.Many real-world settings, however, require reasoning over multiple complementary cues. A song, for example, emerges from the interplay between melody and lyrics, a 3D design may be defined jointly by a sketch and a text description, and a scene can be better understood when visual and auditory signals are considered together. 
Beyond this limitation, multimodal models often exhibit modality competition~\cite{huang2022modality}, where one modality dominates and suppresses others when signal strength or reliability differs. For instance, strong visual cues may overshadow weaker auditory ones, biasing the joint representation toward a single source. Yet many real-world settings require reasoning over complementary cues: a song emerges from melody and lyrics, a 3D design from sketch and text, and a scene from both visual and auditory signals. These challenges motivate a broader question:

% These challenges motivate a broader question:

% \textit{Can a contrastive learning framework capture not only pairwise alignments but also higher-order, synergistic dependencies among modalities?}

\begin{figure}[t]
    \centering
\includegraphics[width=\columnwidth]{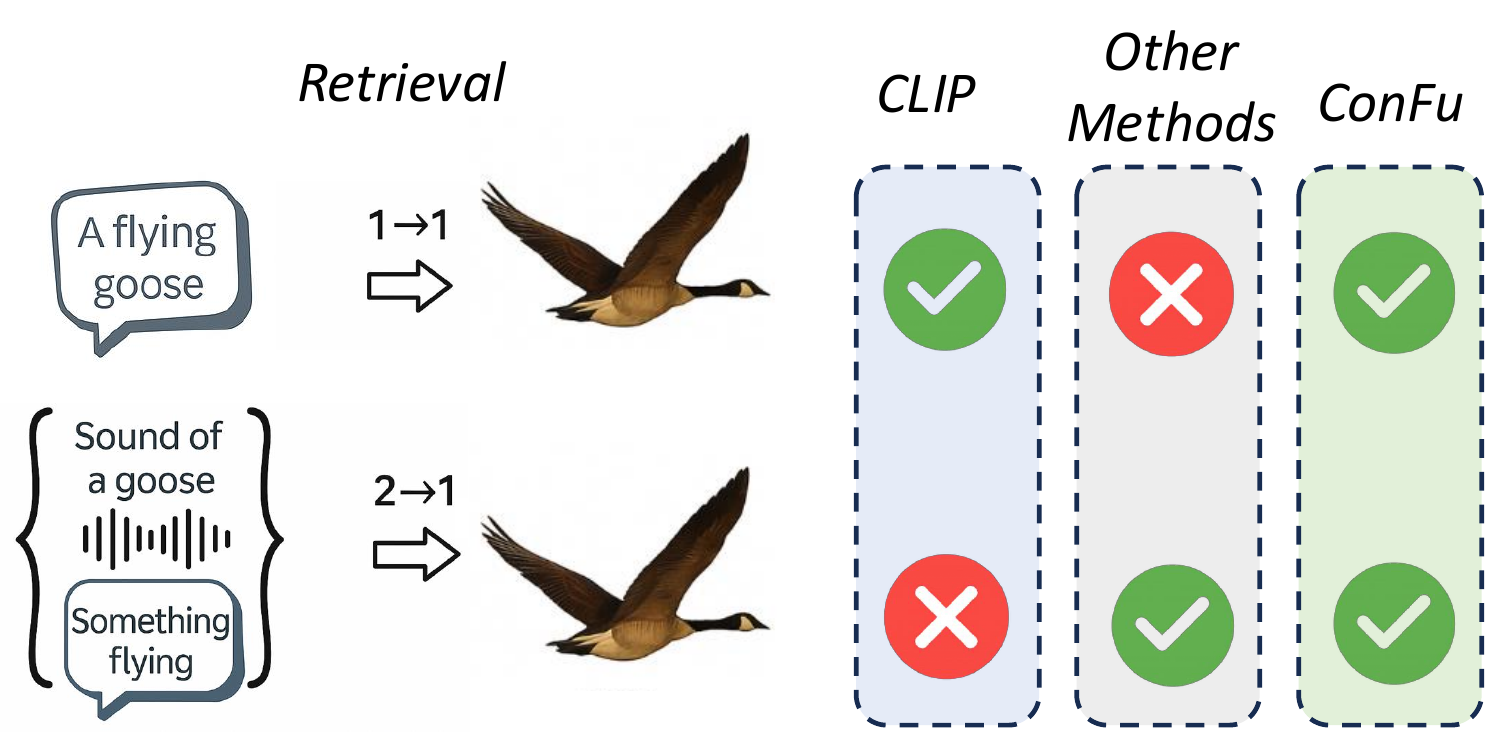}
    \caption{
    ConFu unifies direct (1$\rightarrow$1) and compositional (2$\rightarrow$1) alignment within a single embedding space, utilizing two modalities for improved performance and adapting seamlessly when only one modality is available.
    }
    \label{fig:concept_image}
\end{figure}

\begin{tcolorbox}[
  colback=lightbluebox,
  colframe=lightbluebox,
]
Can a contrastive learning framework capture not only pairwise alignments but also higher-order, synergistic dependencies among modalities?
\end{tcolorbox}

To address this, we propose Contrastive Fusion (ConFu), a framework that unifies pairwise and higher-order multimodal supervision within a single contrastive objective. ConFu constructs fused representations of all subsets of modalities and aligns all combinations of unimodal and fused representations (Fig.~\ref{fig:whole-arch}). 
Unlike previous methods that primarily target specific one-to-one (1$\rightarrow$1)~\cite{guzhov2022audioclip, girdhar2023imagebind}, two-to-one (2$\rightarrow$1) ~\cite{ saporta2024contrasting, cicchetti2025triangle}, or both within predominantly text-centered objectives~\cite{cicchetti2024gramian, chen2023vast}, ConFu accounts for all pairwise and higher-order modality combinations within a unified objective (Fig.~\ref{fig:concept_image}). 

% ~\cite{guzhov2022audioclip, girdhar2023imagebind, saporta2024contrasting, cicchetti2025triangle, chen2023valor, chen2023vast}

% Unlike previous methods focused on one-to-one (1$\rightarrow$1) or two-to-one (2$\rightarrow$1) alignments, ConFu accounts for all pairwise and higher-order combinations within a unified objective. %The framework extends standard contrastive pipelines with minimal overhead while preserving scalability, resulting in richer and more expressive multimodal embeddings that support both 1$\rightarrow$1 and 2$\rightarrow$1 retrieval within a single embedding space.

Evaluating such models is further complicated by the scarcity of publicly available datasets that naturally include three or more co-occurring modalities. To address this, we construct a bird-centric dataset where image, audio, and text provide complementary but related information. This setting enables a systematic examination of higher-order alignment under zero-shot and few-shot regimes.
% Evaluating higher-order multimodal frameworks remains challenging given the scarcity of datasets encompassing more than two modalities. To address this limitation, we extend existing pairwise benchmarks to triplet configurations. Specifically, we construct a bird-centric multimodal dataset where both audio and visual signals provide complementary information, forming artificial triplets that combine image, audio, and textual annotations. This setup enables pretraining and systematic evaluation of models under zero-shot and few-shot conditions, assessing their ability to learn transferable and semantically consistent representations across modalities. We then evaluate the pretrained models on existing real-world bird classification benchmarks to measure generalization and representation quality. Across these tasks, our approach demonstrates competitive or superior performance compared to strong baselines, particularly as multimodal complexity increases.
\noindent Overall, our main contributions are as follows:
\begin{itemize}
\item We introduce a unified contrastive framework that generalizes CLIP-like pairwise multimodal alignment to capture higher-order dependencies among three modalities.
% \item We provide theoretical insight showing that fused supervision yields a stronger and more informative learning signal, promoting the integration of complementary information across modalities.
% \item We develop new synthetic and real-world evaluation suites to assess higher-order multimodal reasoning, addressing the scarcity of existing datasets for this setting.
\item We construct Bird-MML, a synthetic dataset of artificial triplets that serves as a pretraining resource to evaluate whether models capture multimodal complementarity. %, thereby addressing the scarcity of datasets for this setting.
\item We demonstrate that ConFu achieves competitive performance in retrieval and classification tasks, supporting both one-to-one and two-to-one retrieval within a single contrastive framework.
\item We show that ConFu exhibits greater robustness than previous methods when faced with distracting modalities (Table~\ref{tab:zero_shot_results_compact}) and noise-induced distribution shifts (Table~\ref{tab:accuracy_results_noise}), showing greater stability even when fusion involves corrupted or non-informative inputs.
\end{itemize}

\section{Related Work}
% Most contrastive frameworks in multimodal learning operate on paired modalities. Recent extensions, categorized as jointly-aligned or pivot-aligned, expand this formulation to multiple modalities, yet they model higher-order relations only implicitly through shared objectives.
% \subsection{Scaling to Multiple Modalities}

% \paragraph{Jointly-aligned models.}
% Extensions of CLIP to multiple modalities typically train on datasets containing all signals simultaneously. Fusion-based approaches differ in how they combine modalities. {AudioCLIP}~\cite{guzhov2022audioclip} aligns each modality pair independently, while {VALOR}~\cite{chen2023valor} introduces a Multimodal Grouping Alignment loss combined with caption generation, though audio–image interactions remain implicit. {VAST}~\cite{chen2023vast} unifies video, audio, and subtitles through contrastive and generative objectives, forming a shared omni-modal representation. {mPLUG-2}~\cite{xu2023mplug} and {UMT}~\cite{liu2022umt} jointly model vision, audio, and text for understanding and generation tasks using modular decoders, while the {Everything-at-Once Fusion Transformer}~\cite{lim2021temporal} learns a joint embedding space of vision, audio, and text for text-to-audio-visual tasks.

% \paragraph{Jointly-aligned models.}
\textbf{Jointly-aligned models.} Methods such as AudioCLIP~\cite{guzhov2022audioclip}, VALOR~\cite{chen2023valor}, and VAST~\cite{chen2023vast} extend CLIP-style contrastive learning to three or more modalities by training on datasets where all signals co-occur. AudioCLIP employs independent contrastive objectives for each modality pair, while VALOR introduces a multimodal grouping alignment loss coupled with caption generation to connect multiple modalities through text. VAST combines contrastive and generative losses to unify video, audio, and subtitles in a shared representation. Other architectures such as mPLUG-2~\cite{xu2023mplug}, UMT~\cite{liu2022umt}, and the Everything-at-Once Fusion Transformer~\cite{lim2021temporal} employ modular decoders or fusion transformers for multimodal understanding and generation.
While these methods learn representations across multiple modalities, their objectives are typically limited to pairwise alignments or remain primarily text-centric, with broader multimodal relationships emerging implicitly from shared training signals rather than through explicit higher-order alignment.
% Although these approaches learn representations spanning multiple modalities, their objectives optimize only pairwise correspondences, any relationships among larger modality subsets arise implicitly from shared training signals rather than from explicit higher-order alignment.
% Although these models achieve strong multimodal fusion, the underlying pairwise and higher-order relationships are not explicitly modeled but instead emerge indirectly through training. This makes it difficult to assess whether they capture genuine synergistic dependencies among modalities.

\textbf{Pivot-aligned models.}
% An alternative to full multimodal co-occurrence is to use a \emph{pivot} modality that serves as a common reference for alignment. In this paradigm, each non-pivot modality is mapped to the same anchor space, enabling scalable training without requiring all modality combinations to be observed jointly. For example, {ImageBind}~\cite{girdhar2023imagebind} aligns diverse modalities such as audio, depth, and IMU data through a shared image pivot, while {LanguageBind}~\cite{zhu2023languagebind} employs text as the central alignment hub. Building on this idea, recent approaches such as {OmniBind}~\cite{wang2024omnibind}, {FreeBind}~\cite{wang2024freebind}, and {Ex-MCR}~\cite{zhang2024extending} leverage pseudo-pair generation to extend the pivot-based paradigm and further reduce data requirements. Methods such as {Uni3D}~\cite{zhou2023uni3d}, {ULIP}~\cite{xue2023ulip}, and {ULIP-2}~\cite{xue2024ulip} extend CLIP by projecting additional modalities into its frozen embedding space. This pivot-based paradigm improves scalability and data efficiency but remains inherently pairwise. However, these strategies inherently limit cross-modal interactions to those mediated by the pivot, preventing direct modeling of higher-order dependencies among non-pivot modalities. Consequently, the resulting embedding space is formed from independent pairwise relations, limiting the modeling of true higher-order interactions.
An alternative strategy uses a pivot modality as a common reference space. Each non-pivot modality is aligned to this anchor, enabling scalable training without requiring full multimodal co-occurrence. For example, ImageBind~\cite{girdhar2023imagebind} aligns diverse modalities (audio, depth, IMU) to an image-based embedding space, while LanguageBind~\cite{zhu2023languagebind} uses text as the pivot. Later variants such as OmniBind~\cite{wang2024omnibind}, FreeBind~\cite{wang2024freebind}, and Ex-MCR~\cite{zhang2024extending} extend this paradigm through pseudo-pair generation, offering an alternate data-efficient mechanism. Similarly, Uni3D~\cite{zhou2023uni3d}, ULIP~\cite{xue2023ulip}, and ULIP-2~\cite{xue2024ulip} project new modalities into a frozen CLIP embedding space,mapping new modalities to CLIP’s pretrained space without updating its parameters.
While this pivot-based design improves scalability, it remains inherently pairwise: all interactions are mediated through the pivot, preventing direct modeling of dependencies among non-pivot modalities. As a result, the learned embedding space captures independent pairwise relations rather than true higher-order structure.

\textbf{Modeling Higher-Order Dependencies.}
Recent work in self-supervised representation learning has attempted to model relationships between more than two modalities through specialized loss formulations. Symile~\cite{saporta2024contrasting} leverages total correlation to integrate complementary cues, TRIANGLE~\cite{cicchetti2025triangle} uses the area spanned by three embedding vectors as a measure of similarity, and GRAM~\cite{cicchetti2024gramian} measures cross-modal similarity for $M \ge 2$ modalities through the Gramian volume spanned by their vectors. While these methods jointly model multiple modalities, Symile and TRIANGLE require all modalities to be present at inference time, making them incompatible with standard 1$\rightarrow$1 retrieval. 

% Moreover, geometric formulations that minimize areas or volumes behave similarly to cosine similarity, which may limit their ability to capture complex inter-modal relations.

In contrast, our framework explicitly models both pairwise and higher-order alignments within a single contrastive objective. It provides a simple, architecture-agnostic formulation that unifies pairwise and higher-order multimodal interactions, enabling scalable and expressive cross-modal representation learning, with the only additional computational cost coming from lightweight MLP layers.

\section{Method}

\subsection{Proposed Multimodal Contrastive Framework}
\label{sec:method}

\begin{figure}
    \centering
    \includegraphics[width=\linewidth]{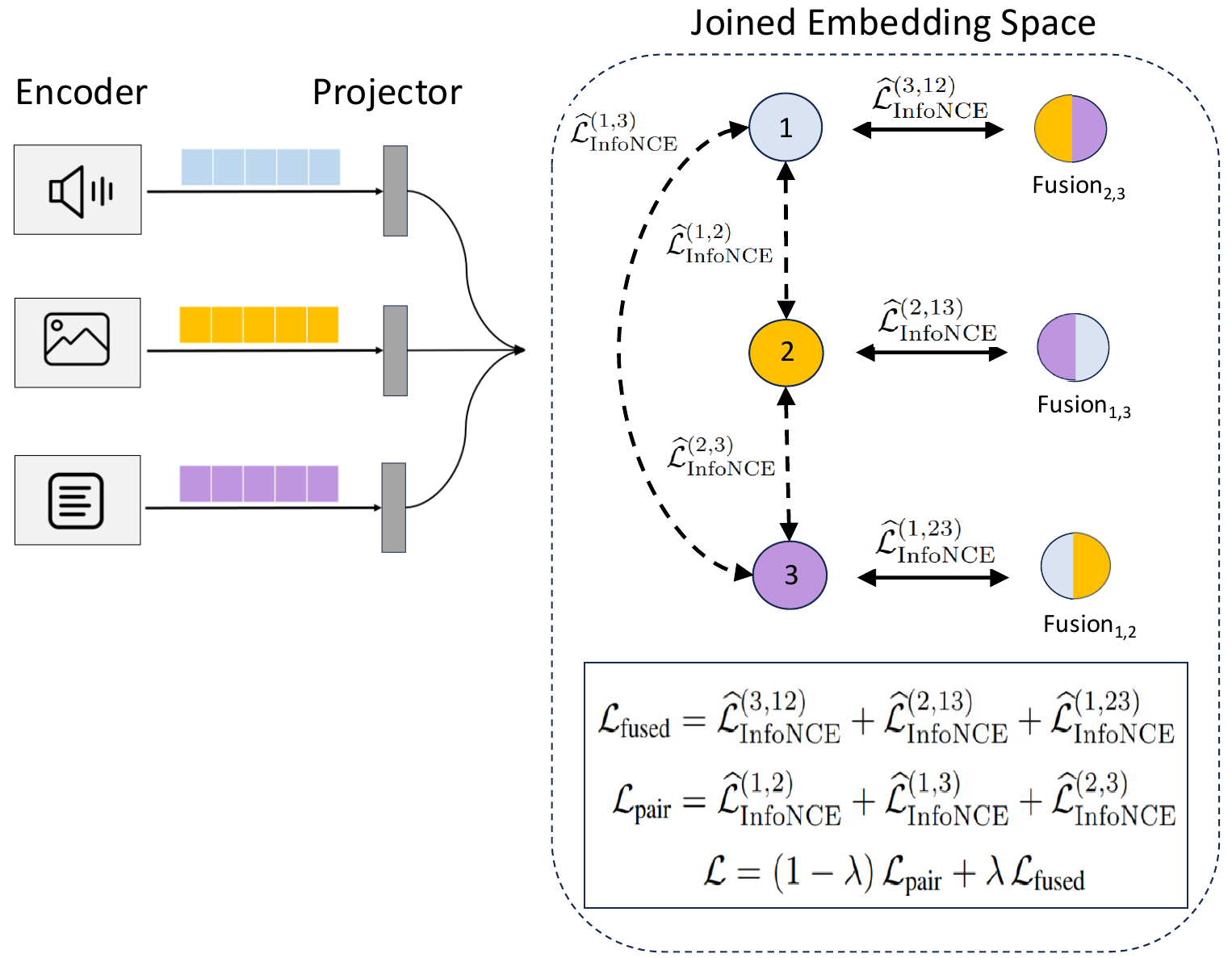}
    \caption{Overview of ConFu. The framework aligns all modality pairs through pairwise contrastive objectives while also aligning each modality with the fused representation of the remaining ones. The final loss ($\mathcal{L}$) combines both
objectives (${\mathcal{L}_{pair}}$, ${\mathcal{L}_{fused}}$), balanced by a weighting factor $\lambda$.}
    \label{fig:whole-arch}
\end{figure}

We propose a new framework for contrastive learning with $M = 3$ modalities (Fig.~\ref{fig:whole-arch}) that unifies \emph{pairwise} and \emph{higher-order} contrastive objectives into a single multimodal training process that maximizes a lower bound on the total correlation among modalities (Sec.~\ref{sec:tc}).

Each modality $X_i$ is mapped into a feature space by a modality-specific encoder $f_{\theta_i}:\mathcal{X}_i\!\to\!\mathcal{H}_i$ and then projected into a shared latent space $\mathcal{Z}$ by a projector $p_{\phi_i}:\mathcal{H}_i\!\to\!\mathcal{Z}$:
\begin{equation}
h_i = f_{\theta_i}(X_i), 
\quad
z_i = p_{\phi_i}(h_i),
\quad i \in \{1,2,3\}.
\label{eq:enc-proj}
\end{equation}
The encoder parameters $\theta_i$ and projector parameters $\phi_i$ are learned jointly through contrastive optimization.

\paragraph{Pairwise Contrastive Objectives ($1{\rightarrow}1$).}
For each modality pair $(X_i, X_j)$, we estimate their MI $I(X_i; X_j)$ via the InfoNCE lower bound Eq.~\ref{eq:pairwise-loss}.  
Let $s_{\omega_{ij}}:\mathcal{Z}\!\times\!\mathcal{Z}\!\to\!\mathbb{R}$ denote a density ratio function parameterized by $\omega_{ij}$, which models the likelihood of joint co-occurrence over independence. Given a minibatch of $N$ paired samples $\{(z_i^{(m)},z_j^{(m)})\}_{m=1}^N$, the empirical loss is
\begin{equation}
\resizebox{0.9\columnwidth}{!}{%
$
\widehat{\mathcal{L}}_{\mathrm{InfoNCE}}^{(i,j)}
=
\frac{1}{N}\!\sum_{m=1}^{N}
-\log
\frac{
\exp\!\{s_{\omega_{ij}}(z_i^{(m)},z_j^{(m)})\}
}{
\sum_{n=1}^{N}\exp\!\{s_{\omega_{ij}}(z_i^{(m)},z_j^{(n)})\}
}.$}
\label{eq:pairwise-loss}
\end{equation}
Summing over all ordered modality pairs yields
\begin{equation}
\mathcal{L}_{\text{pair}}
=
\widehat{\mathcal{L}}_{\mathrm{InfoNCE}}^{(1,2)}
+
\widehat{\mathcal{L}}_{\mathrm{InfoNCE}}^{(1,3)}
+
\widehat{\mathcal{L}}_{\mathrm{InfoNCE}}^{(2,3)}.
\label{eq:pair-loss-total}
\end{equation}

\paragraph{Higher-Order Contrastive Objectives ($2{\rightarrow}1$).}
To model higher-order dependencies $I(X_k; X_i,X_j)$, we introduce a fusion network
$g_{\psi_{ij}}:\mathcal{H}_i\!\times\!\mathcal{H}_j\!\to\!\mathcal{Z}$
with encoded features:
\begin{equation}
z_{ij} = g_{\psi_{ij}}(h_i, h_j),
\quad (i,j)\!\in\!\{(1,2),(1,3),(2,3)\},
\label{eq:fusion}
\end{equation}
where $\psi_{ij}$ are trainable fusion parameters. In our experiments, we have maintained $g$ to be a small-scale network (e.g., a shallow MLP) to capture cross-modal interactions without significantly increasing model complexity. Each fused representation $z_{ij}$ is then aligned with the remaining modality $X_k$, using the ratio density estimation function
$t_{\eta_{k,ij}}:\mathcal{Z}\!\times\!\mathcal{Z}\!\to\!\mathbb{R}$ 
parameterized by $\eta_{k,ij}$:
\begin{equation}
\resizebox{0.9\columnwidth}{!}{%
$
\widehat{\mathcal{L}}_{\mathrm{InfoNCE}}^{(k,ij)}
=
\frac{1}{N}\!\sum_{m=1}^{N}
-\log
\frac{
\exp\!\{t_{\eta_{k,ij}}(z_k^{(m)},z_{ij}^{(m)})\}
}{
\sum_{n=1}^{N}\exp\!\{t_{\eta_{k,ij}}(z_k^{(m)},z_{ij}^{(n)})\}
}.$}
\label{eq:fused-loss}
\end{equation}
In practice, both density ratio estimators $s$ in Eq.~\ref{eq:pairwise-loss} and $t$ in Eq.~\ref{eq:fused-loss} are implemented as temp.-scaled dot-product similarities. The higher-order objective aggregates all such triplets:
\begin{equation}
\mathcal{L}_{\text{fused}}
=
\widehat{\mathcal{L}}_{\mathrm{InfoNCE}}^{(3,\{1,2\})}
+
\widehat{\mathcal{L}}_{\mathrm{InfoNCE}}^{(2,\{1,3\})}
+
\widehat{\mathcal{L}}_{\mathrm{InfoNCE}}^{(1,\{2,3\})}.
\label{eq:fused-loss-total}
\end{equation}

% {\color{blue}\paragraph{Why higher order?}
% By monotonicity of mutual information with respect to its arguments,
% \begin{equation}
% I([X_i,X_j]; X_k) \;\ge\; \max\{\,I(X_i; X_k),\, I(X_j; X_k)\,\},
% \label{eq:mi-monotonicity}
% \end{equation}
% where $[X_i,X_j]$ denotes any form of fusion. Thus, aligning the fused pair $[X_i,X_j]$ with $X_k$ is at least as informative as any single pairwise alignment, providing an additional justification for our higher-order objective. A formal proof of this inequality is given in Appendix~\ref{}.}

\paragraph{Combined Objective.}
The total training loss jointly maximizes both pairwise and fused MI lower bounds:
\begin{equation}
\mathcal{L}
= (1-\lambda)\,\mathcal{L}_{\text{pair}}
+ \lambda\,\mathcal{L}_{\text{fused}},
\label{eq:final-loss}
\end{equation}
where $\lambda\!\in\![0,1]$ controls the relative contribution of pairwise versus higher-order supervision (See Appendix~\ref{appendix:ablation_lambda} for analysis of $\lambda$’s impact).
Minimizing Eq.~\ref{eq:final-loss} corresponds to maximizing the contrastive lower bound on $\mathrm{TC}(X_1,X_2,X_3)$ (Eq.~\ref{eq:tc-infonce}), thus promoting both (i) pairwise alignment and (ii) cross-modal consistency through fused interactions.

\subsection{Theoretical Motivation}
\label{sec:tc}

The total correlation (TC)~\citep{watanabe1960information}, also known as multi-information~\citep{studeny1998multiinformation}, quantifies the joint statistical dependency among multiple random variables as the Kullback–Leibler divergence between their joint distribution and the product of marginals,  
$\mathrm{TC}(X_1,X_2,X_3) = D_{\mathrm{KL}}\!\big(p(x_1,x_2,x_3)\,\|\,p(x_1)p(x_2)p(x_3)\big)$.  
Equivalently,
\begin{equation}
\resizebox{0.9\columnwidth}{!}{$
\mathrm{TC}(X_1,X_2,X_3)
= 
\mathbb{E}_{\!\!p(x_1,x_2,x_3)}
\!\left[\log
\frac{p(x_1,x_2,x_3)}{p(x_1)p(x_2)p(x_3)}\right],
\label{eq:tc-def}
$}
\end{equation}
and vanishes if and only if the variables are mutually independent ($X_1 \perp X_2 \perp X_3$).
 Using the chain rule of MI, it decomposes as
\begin{equation}
\mathrm{TC}(X_1,X_2,X_3)
= I(X_1;X_2) + I(X_3;X_1,X_2).
\label{eq:tc-decomp}
\end{equation}
Averaging over all permutations yields the symmetric form
\begin{equation}
\resizebox{0.88\columnwidth}{!}{$
\mathrm{TC}(X_1,X_2,X_3)
= \tfrac{1}{3}\! \hspace{-20pt}
\underset{(i,j,k)\in\text{perm}\{1,2,3\}}{\sum} \hspace{-20pt}
\!\big[I(X_i;X_j) + I(X_k;X_i,X_j)\big],
$}
\label{eq:tc-symmetric}
\end{equation}
where $\text{perm}\{1,2,3\}$ denotes the set of all six permutations of the index triplet $(1,2,3)$. This interpretation of TC separates pairwise and higher-order dependencies.

\paragraph{Contrastive Lower Bound.}
Each MI term in Eq.~\ref{eq:tc-symmetric} is approximated using the InfoNCE lower bounds~\cite{oord2018representation}.  
Specifically, the pairwise and higher-order terms satisfy
\begin{align}
I(X_i;X_j)
&\;\ge\;
\log N - \widehat{\mathcal{L}}_{\mathrm{InfoNCE}}^{(i,j)},
\label{eq:infonce-bound-pair} \\
I(X_k;X_i,X_j)
&\;\ge\;
\log N - \widehat{\mathcal{L}}_{\mathrm{InfoNCE}}^{(k,ij)}.
\label{eq:infonce-bound-triplet}
\end{align}
Substituting these inequalities into Eq.~\ref{eq:tc-symmetric} yields the tractable contrastive lower bound on $\mathrm{TC}$, 
% \begin{equation}
% \resizebox{\columnwidth}{!}{
% \begin{align}
% \mathrm{TC}(X_1,X_2,X_3)
% &\ge
% - \tfrac{1}{3}\!\hspace{-20pt}
% \underset{(i,j,k)\in\text{perm}\{1,2,3\}}{\sum} \hspace{-20pt}\!
% \Big[  \widehat{\mathcal{L}}_{\mathrm{InfoNCE}}^{(i,j)}
% + \widehat{\mathcal{L}}_{\mathrm{InfoNCE}}^{(k,ij)}
% \Big]
% + 2 \log N + \text{const.}
% \end{align}
% }
% \label{eq:tc-infonce}
% \end{equation}
\begin{equation}
\scalebox{0.92}{%
\(
\begin{aligned}
\mathrm{TC}
&\ge
-\tfrac{1}{3}\!\hspace{-20pt}
\sum_{(i,j,k)\in\text{perm}\{1,2,3\}} \hspace{-20pt}
\Big[
\widehat{\mathcal{L}}_{\mathrm{InfoNCE}}^{(i,j)}
+
\widehat{\mathcal{L}}_{\mathrm{InfoNCE}}^{(k,ij)}
\Big]
+ 2\log N + \text{const.}
\end{aligned}
\)
}
\label{eq:tc-infonce}
\end{equation}
so that maximizing total correlation reduces to minimizing the InfoNCE losses across all pairwise and higher-order relations.  
When each density ratio function $s_{\omega_{ij}}$ or $t_{\eta_{k,ij}}$ is sufficiently expressive and approximates the optimal InfoNCE critic~\cite{oord2018representation} (i.e., the log–density ratio between the joint and the product of marginals) the bound becomes tight as $N\!\to\!\infty$, recovering the true $\mathrm{TC}$.

\paragraph{Interpretation.}
Eq.~\ref{eq:tc-infonce} shows that the proposed training objective (Eq.~\ref{eq:final-loss}) maximizes a contrastive lower bound on total correlation, thereby promoting both (i) pairwise alignment across modalities through $I(X_i;X_j)$ and (ii) higher-order consistency via $I(X_k;X_i,X_j)$. 

% By an information-theoretic argument, 
% \begin{equation}
% I([X,Y]; Z) \ge \max\{I(X; Z), I(Y; Z)\},
% \end{equation}
% where $[X,Y]$ represents the concatenated (or fused) modalities. 

% This inequality implies that aligning the fused pair $[X,Y]$ with $Z$ 
% captures at least as much mutual information as aligning any single pair 
% $(X,Z)$ or $(Y,Z)$. 
% Consequently, contrastive objectives defined over fused pairs 
% provide stronger supervision, encouraging the model to exploit 
% joint multimodal cues. 

\subsection{Compare with other higher-order methods}

% Recent approaches such as SYMILE, GRAM, and TRIANGLE optimize a \emph{single joint alignment objective} across all modalities simultaneously, without disentangling pairwise dependencies. SYMILE~\citep{saporta2024contrasting} formulates this as a total correlation bound over all embeddings, $\mathcal{L}_{\operatorname{SYMILE}} = -\mathbb{E}\!\left[ \log \frac{\exp(\langle x_1, x_2, x_3\rangle / \tau)} {\sum_j \exp(\langle x_1, x_2, x_3^{(j)}\rangle / \tau)} \right]$, where the multilinear inner product $\langle x_1, x_2, x_3\rangle$ captures joint statistical dependence among modalities. GRAM~\citep{cicchetti2024gramian} adopts a geometric perspective, enforcing global alignment by minimizing the parallelotope volume $V(x_1, x_2, x_3) = \sqrt{\det G(x_1, x_2, x_3)}$, while TRIANGLE~\citep{cicchetti2025triangle} specializes this idea to three modalities through the triangle area $A(x_1, x_2, x_3) = \frac{1}{2}\sqrt{ \langle u, u\rangle \langle v, v\rangle - \langle u, v\rangle^2 }$, with $u = x_1 - x_2$ and $v = x_1 - x_3$, both measuring the overall co-alignment of $(x_1, x_2, x_3)$ in latent space.

Recent approaches such as Symile, GRAM, and TRIANGLE adopt a \emph{single joint alignment objective} across all modalities, without separating pairwise from higher-order dependencies.
Symile maximizes a total-correlation bound using a multilinear similarity $\langle x_1, x_2, x_3\rangle$,
GRAM enforces global alignment via the parallelotope volume induced by the Gram matrix,
and TRIANGLE specializes this idea by introducing a triangle-area–based measure of three-way similarity.
In contrast, our method decomposes total correlation into interpretable information-theoretic factors, averaging over modality permutations the sum of pairwise and higher-order dependencies, thus preserving holistic expressivity while enabling structured optimization and reliable performance with any subset of modalities. Unlike prior approaches that use a single critic to implicitly model dependencies, we factorize them at the loss level, enabling independent control of pairwise and higher-order supervision while preserving InfoNCE stability and achieving balanced performance within a single model.
\subsection{Motivating higher-order dependencies}

\begin{figure}[t]
    \centering

    \includegraphics[width=\linewidth]{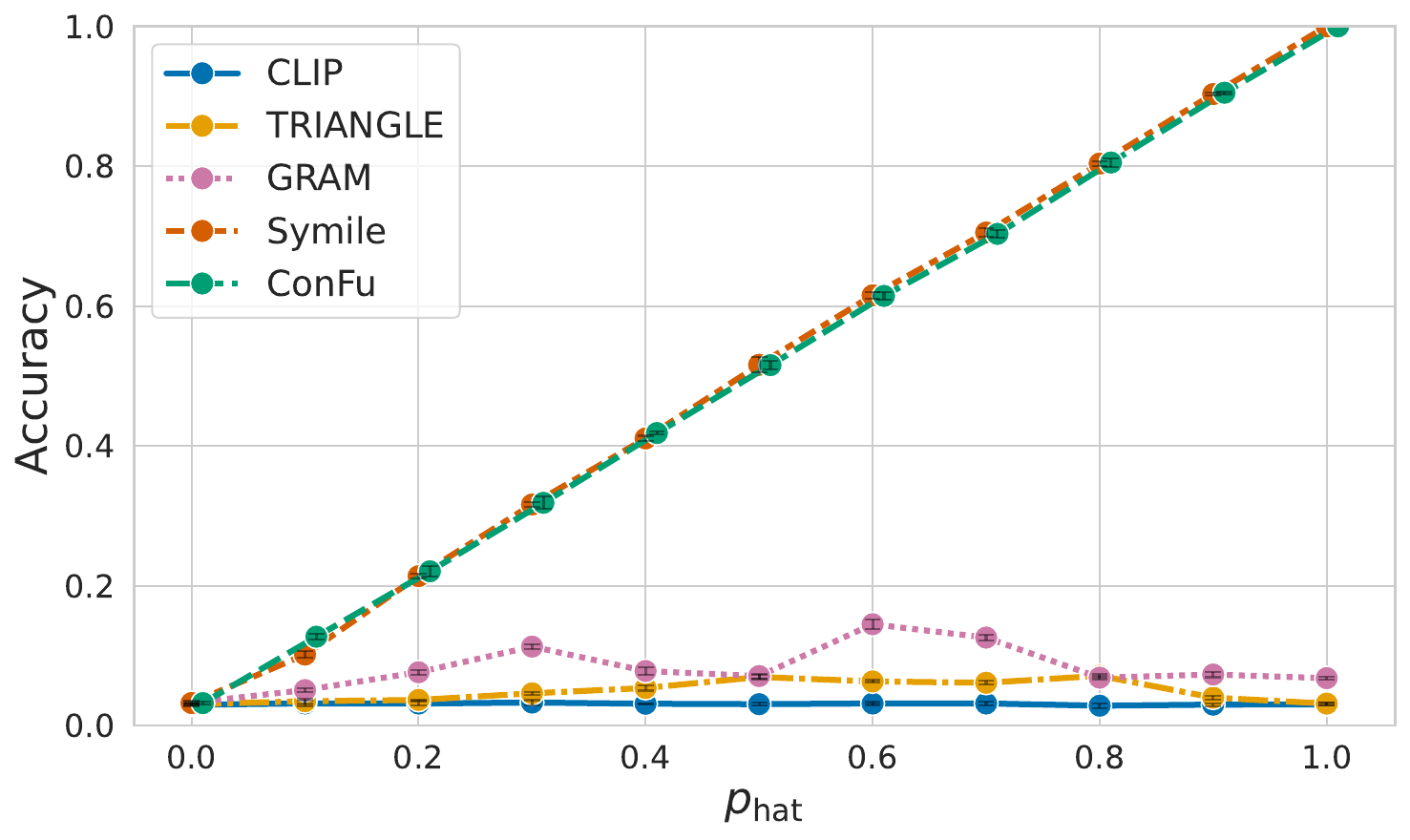}
    \caption{
    Results on the synthetic XOR task. Our model's accuracy in predicting $z_2$ from $(z_1, z_3)$ is plotted against the mixing parameter $\hat{p}$. Our model captures the synergistic information, showing a positive trend in performance as $\hat{p}$ increases. Trimodal pairwise CLIP remains on chance ($\sim3\%$) while both GRAM and TRIANGLE fail to reach above 15\% accuracy. More details are provided in Appendix \ref{appendix:ablation_xor}. 
    }
    \label{fig:synthetic_xor}
\end{figure}

To illustrate the reasons that modeling higher-order (synergistic) dependencies is crucial, 
we consider the {XOR task} introduced in Symile. Three binary variables $X_1$, $X_2$, and $X_3$ are sampled with ${x_1}_j,{x_2}_j \sim \mathrm{Bernoulli}(0.5)$,
$i \sim \mathrm{Bernoulli}(\hat{p})$, and the relationship:
\begin{equation}
    {x_3}_j = ({x_1}_j \oplus {x_2}_j)^i {x_1}_j^{(1-i)}.
\end{equation}
The task is to predict the representation ${z_2}$ from the pair $({z_1}, {z_3})$. The parameter $\hat{p}$ controls the level of synergy: for $\hat{p}=0$, ${x_2}$ is independent of $({x_1}, {x_3})$, while increasing $\hat{p}$ raises the conditional MI $I(X_1;X_2|X_3)$ and $I(X_3;X_2|X_1)$ without introducing any pairwise dependence ($I(X_1;X_2)=I(X_2;X_3)=0$). Thus, only models capturing joint interactions, rather than pairwise correlations, can solve the task. 
Zero-shot retrieval at varying levels of $\hat{p}$ (Fig.~\ref{fig:synthetic_xor}) confirms that ConFu and Symile can successfully solve the task.

\section{Bird-MML: A Multimodal Dataset for Audio-Visual Complementarity}
% Existing datasets provide valuable resources but have notable limitations for multimodal pretraining. However, there is currently no standardized dataset that jointly combines images, audio, and text in a form suitable for multimodal pretraining and evaluation. 

Despite substantial progress in multimodal learning, there remains a lack of standardized and publicly available realistic datasets that capture cross-modal relationships that are suitable for evaluating both pairwise and higher-order multimodal alignment. %Ιnstead, existing resources are typically limited to two modalities and are not designed with higher-order information in mind.
Bird species provide an ideal testbed for such multimodal evaluation, as their identity and behavior are naturally expressed through both visual and acoustic cues. In many cases, fine-grained recognition further depends on complementary information such as visual appearance for morphology and audio for vocalization, making the domain well suited for studying multimodal interaction.

\textbf{Existing Bird Datasets.} SSW60~\cite{van2022exploring} offers unaligned images and audio as well as video for 60 species, VB100~\cite{ge2016exploiting} focuses on fine-grained video recognition for 100 species. Moreover, iNatSounds~\cite{chasmai2024inaturalist} contains large-scale bioacoustic recordings without visual counterparts, and lastly, BioTrove-Train~\cite{yang2024biotrove} includes millions of images across diverse species with uneven per-class distributions. Although those datasets provide valuable resources, a joint combination of images, audio, and text in a form suitable for multimodal pretraining and evaluation is missing.

\textbf{Dataset Overview.}
To bridge this gap, we construct Bird-MML, a synthetic multimodal bird dataset integrating artificially paired images, audio, and text for 150 species, including all those present in SSW60 and VB100. The dataset is designed for scalable pretraining and cross-modal learning, with balanced representation across species. A detailed overview of the data generation pipeline is provided in Appendix~\ref{appendix:datasets_bird_mml}. Images are sourced from the iNaturalist open dataset, restricted to research-grade, Creative Commons–licensed observations. For each species, 1,000 samples are randomly selected.
% To bridge this gap, we construct a multimodal bird dataset integrating artificially paired images, audio, and text for 150 species,  including all those present in SSW60 and VB100. The dataset is designed for scalable pretraining and cross-modal learning, with balanced representation across species. A detailed overview of the data generation pipeline is provided in Appendix ~\ref{appendix:datasets}. Images are sourced from the iNaturalist open dataset, restricted to research-grade, Creative Commons–licensed observations. For each species, 1,000 samples are randomly selected. 

\textbf{Audio and Text Generation.} Audio recordings are collected from Xeno-Canto~\cite{xenocanto}, segmented into 10-second clips, and zero-padded when necessary. In cases of limited availability, existing clips are reused to maintain uniform sample counts. Metadata such as call type, sex, and life stage are preserved. Text descriptions are generated by combining three complementary sources: image captions from InstructBLIP2~\cite{dai2023instructblip}, audio metadata, and short summaries from corresponding Wikipedia entries. These elements are fused using \texttt{google/gemma-2-2b-it}~\cite{gemma_2024} to yield a single multimodal caption per instance. 

\textbf{Dataset Characteristics.} While generated captions may contain minor factual noise, they effectively capture species-relevant attributes and contextual cues useful for representation learning. The resulting dataset contains 149,681 triplets (image, audio, text), each grounded by species identity and visual–acoustic correspondence. Approximately 43\% of species required limited audio reuse due to data scarcity. Examples of artificially aligned triplets are shown in Appendix~\ref{appendix:datasets_bird_mml}.The dataset maintains balanced per-species coverage and is well-suited for multimodal pretraining. Downstream evaluations are performed on naturally paired video-audio data to assess real-world cross-modal performance. 

\textbf{Ethical Considerations.} All data were collected from publicly available sources under Creative Commons licenses. Ownership and attribution remain with original contributors. Automatically generated captions may contain minor factual errors or biases and should not be interpreted as verified biological information. The dataset is released solely for non-commercial research use.

\newcommand{\sres}[2]{#1 {\scriptsize$\pm$ #2}} % smaller std only

\begin{table}[t]
\caption{
Zero-shot classification accuracy on AV-MNIST (mean $\pm$ std over 5 runs).  
A = Audio, V = Vision, A+V = Audio–Visual fusion. Best performance is reported in bold.
}
\centering
\begin{tabular}{lccc}
\toprule
\textbf{Method} & \textbf{A} & \textbf{V} & \textbf{A+V} \\
\midrule
CLIP & \textbf{\sres{41.1}{0.3}} & \sres{63.0}{0.4} & - \\
Tri-CLIP & \sres{40.0}{0.4} & \sres{62.5}{0.4} & - \\
Symile ~\cite{saporta2024contrasting} & -  & - & \sres{70.9}{0.3} \\
GRAM ~\cite{cicchetti2024gramian} & \sres{9.8}{0.4} & \sres{63.9}{0.8} & \sres{64.4}{0.5} \\
TRIANGLE ~\cite{cicchetti2025triangle} & - & - & \sres{64.9}{0.3} \\
\textbf{ConFu} & \sres{39.7}{0.1} & \textbf{\sres{64.6}{0.3}} & \textbf{\sres{71.2}{0.2}} \\
\bottomrule
\end{tabular}
\label{tab:avmnist_results}
\end{table}

\section{Experimental Results}
\label{sec:results}
Our evaluation covers multimodal benchmarks across biodiversity ({SSW60}~\cite{van2022exploring}, {VB100}~\cite{ge2016exploiting}), affective ({MOSI}~\cite{zadeh2016mosi}, {UR-FUNNY}~\cite{hasan2019ur}, and {MUStARD}~\cite{castro2019towards} included in {MultiBench}~\cite{liang2021multibench}), and multimedia ({AV-MNIST}~\cite{vielzeuf2018centralnet}) domains, encompassing both real-world and synthetic datasets.

% We evaluate our model across diverse multimodal benchmarks spanning synthetic, affective, and real-world domains, including datasets from {MultiBench}~\cite{liang2021multibench}, {AV-MNIST}~\cite{vielzeuf2018centralnet}, and fine-grained bird benchmarks {SSW60}~\cite{van2022exploring} and {VB100}~\cite{ge2016exploiting}.
These datasets present different levels of cross-modal difficulty, enabling holistic evaluation of alignment, representation quality, and scalability.
We compare against representative contrastive and higher-order multimodal approaches:
(i) a {Bi-modal CLIP} (referenced as CLIP from now on) trained with standard two-modality contrastive loss,
(ii) a {Tri-modal CLIP} (referenced as Tri-CLIP from now on) trained with pairwise objectives across all modalities,
(iii) {Symile}~\cite{saporta2024contrasting} which models higher-order dependencies via total correlation,
(iv) {TRIANGLE}~\cite{cicchetti2025triangle} and {GRAM}~\cite{cicchetti2024gramian}, which use geometric similarity measures among 2 or more modalities.
Experiments cover (i) zero-shot classification, (ii) 1$\rightarrow$1 and 2$\rightarrow$1 retrieval, and (iii) linear-probe classification under few-shot and full-data regimes. Details about the experimental setup and evaluation can be found in Appendix \ref{appendix:experimental_details}.

\subsection{AV-MNIST}

{We evaluate our framework on the AV-MNIST dataset, which contains paired audio spectrograms augmented with randomly sampled background noise from ESC-50 ~\cite{piczak2015dataset} and images of MNIST digits with degraded visual features. We additionally introduce a text modality derived from class labels using natural language templates such as
\textit{`the digit is a \{class\}''}, \textit{`a photo of a \{class\}''}, and \textit{`a picture of a \{class\}''}.
}
% We perform zero-shot classification, where no labeled examples are used for training—the model relies solely on text prompts as semantic descriptors.
As provided in Table~\ref{tab:avmnist_results}, our approach achieves the highest performance across all baselines, with Symile trailing closely behind.
Fusing audio and visual inputs yields an 8\% improvement over the strongest unimodal baseline, demonstrating the benefit of complementary multimodal supervision.
Moreover, even when evaluated on a single modality (vision), our model surpasses the corresponding unimodal baseline by 1.5\%, indicating that multimodal training enhances individual modality representations.
% ConFu also yields consistent results across runs, showing that multimodal fusion stabilizes training and reduces sensitivity to initialization.

% \renewcommand{\arraystretch}{1.1}
% \begin{table}[t]
% \caption{
% Zero-shot classification accuracy on AV-MNIST (mean $\pm$ std over 5 runs).  
% A = Audio, V = Vision, T = Text. Best result in \textbf{bold}, second best \underline{underlined}.
% }
% \centering
% \begin{tabular}{lccc}
% \toprule
% \textbf{Method} & \textbf{A} & \textbf{V} & \textbf{A+V} \\
% \midrule
% CLIP & \sres{41.09}{0.26} & \sres{63.01}{0.44} & - \\
% Tri-CLIP & \sres{40.01}{0.43} & \sres{62.54}{0.43} & - \\
% Symile & - & - & \sres{70.91}{0.30} \\
% \textbf{ConFu} & \sres{39.69}{0.13} & \underline{\sres{64.65}{0.28}} & \textbf{\sres{71.21}{0.22}} \\
% \bottomrule
% \end{tabular}
% \label{tab:avmnist_results}
% \end{table}

% \renewcommand{\arraystretch}{1.0}

\newcommand{\res}[2]{\large #1 {\scriptsize$\pm$ #2}}
\begin{table*}
\caption{Recall@10 (\%) for \textbf{Target} across query modalities and datasets (mean $\pm$ std over 5 runs). The last two columns show mean retrieval accuracy for 1→1 and 2→1 retrieval, respectively. Best performance is reported in bold, second best is underlined.}
\centering
\resizebox{\textwidth}{!}{
\begin{tabular}{ll|ccc|ccc|ccc|cc}
\toprule
 & Target &
\multicolumn{3}{c|}{\textbf{M1}} &
\multicolumn{3}{c|}{\textbf{M2}} &
\multicolumn{3}{c|}{\textbf{M3}} &
\multicolumn{2}{c}{\textbf{Mean}} \\[-2pt]
\toprule
& Query & \textbf{M2} & \textbf{M3} & \textbf{M23} &
  \textbf{M1} & \textbf{M3} & \textbf{M13} &
  \textbf{M1} & \textbf{M2} & \textbf{M12} &
  \textbf{\scriptsize 1→1} & \textbf{\scriptsize 2→1} \\
\midrule
\multirow{5}{*}{\rotatebox{90}{\textbf{MOSI}}}
& Trimodal CLIP & \textbf{\res{22.9}{1.9}} & \textbf{\res{20.5}{1.6}} & -- & \textbf{\res{23.5}{3.6}} & \textbf{\res{23.3}{1.8}} & -- & \textbf{\res{20.9}{2.0}} & \textbf{\res{24.3}{2.1}} & -- & \textbf{22.6} & -- \\
& Symile ~\cite{saporta2024contrasting} & - & - & \underline{\res{16.3}{0.9}} & -  & -  &\underline{\res{18.1}{2.4}} & -  & -  & \underline{\res{17.1}{1.7}} & -- & \underline{17.2} \\
& GRAM~\cite{cicchetti2024gramian} & \res{6.0}{3.1} & \res{10.3}{2.9} & \res{16.3}{2.5} & \res{7.7}{2.3} & \res{0.3}{0.6} & \res{7.9}{3.4} & \res{12.2}{4.1} & \res{0.2}{0.2} & \res{12.0}{4.2} & 6.1 & 12.1 \\
& TRIANGLE~\cite{cicchetti2025triangle} & -  & -  & \res{8.3}{1.3} & -  & -  & \res{4.9}{0.7} & -  & -  & \res{8.8}{1.4} & -- & 7.3 \\
& \textbf{ConFu} & \cellcolor{green!15}\underline{\res{21.0}{1.7}} & \cellcolor{green!15}\underline{\res{16.4}{2.3}} & \cellcolor{green!15}\textbf{\res{16.7}{1.8}} & \cellcolor{green!15}\underline{\res{19.2}{1.4}} & \cellcolor{green!15}\underline{\res{21.0}{2.4}} & \cellcolor{green!15}\textbf{\res{21.6}{1.9}} & \cellcolor{green!15}\underline{\res{16.1}{1.7}} & \cellcolor{green!15}\underline{\res{23.5}{2.7}} & \cellcolor{green!15}\textbf{\res{20.5}{2.9}} & \cellcolor{green!15}\underline{19.5} & \cellcolor{green!15}\textbf{19.6} \\
\midrule
\multirow{5}{*}{\rotatebox{90}{\textbf{UR-FUNNY}}}
& Trimodal CLIP & \textbf{\res{3.7}{0.6}} & \textbf{\res{4.0}{0.6}} & -- & \textbf{\res{3.8}{0.5}} & \textbf{\res{16.2}{1.6}} & -- & \textbf{\res{4.0}{0.4}} & \textbf{\res{16.8}{1.2}} & -- & \textbf{8.1} & -- \\
& Symile ~\cite{saporta2024contrasting} & -  & -  & \res{3.7}{0.5} & -  & -  & \res{15.4}{0.9} & -  & -  & \underline{\res{16.5}{0.6}} & -- & \underline{11.9} \\
& GRAM~\cite{cicchetti2024gramian} & \res{3.2}{0.5} & \res{3.0}{0.9} & \underline{\res{3.9}{0.8}} & \res{3.2}{0.9} & \res{0.1}{0.2} & \res{3.1}{0.4} & \underline{\res{3.9}{0.4}} & \res{0.1}{0.0} & \res{3.3}{0.5} & 2.3 & 3.4 \\
& TRIANGLE~\cite{cicchetti2025triangle} &-  & -  & \textbf{\res{4.2}{0.4}} & -  & -  & \res{3.3}{0.7} & -  & -  & \res{3.6}{0.9} & -- & 3.7 \\
& \textbf{ConFu} & \cellcolor{green!15}\underline{\res{3.2}{0.3}} & \cellcolor{green!15}\underline{\res{3.5}{0.3}} & \res{3.6}{0.3} & \cellcolor{green!15}\underline{\res{3.5}{0.2}} & \cellcolor{green!15}\underline{\res{15.1}{0.8}} & \cellcolor{green!15}\textbf{\res{16.9}{0.9}} & \res{3.5}{0.4} & \cellcolor{green!15}\underline{\res{15.6}{0.8}} & \cellcolor{green!15}\textbf{\res{20.3}{1.1}} & \cellcolor{green!15}\underline{7.4} & \cellcolor{green!15}\textbf{13.6} \\
\midrule
\multirow{5}{*}{\rotatebox{90}{\textbf{MUStARD}}}
& Trimodal CLIP & \underline{\res{70.7}{4.1}} & \underline{\res{29.1}{3.6}} & -- & \underline{\res{70.5}{4.3}} & \underline{\res{26.5}{3.2}} & -- & \underline{\res{30.4}{3.4}} & \underline{\res{24.6}{2.8}} & -- & \underline{42.0} & -- \\
& Symile ~\cite{saporta2024contrasting} & -  & - & \res{61.5}{6.3} & -  & -  & \res{57.0}{3.0} & -  & -  & \res{21.3}{4.4} & -- & 46.6 \\
& GRAM~\cite{cicchetti2024gramian} & \res{61.7}{8.0} & \res{24.8}{5.3} & \textbf{\res{81.0}{4.1}} & \res{67.2}{5.4} & \res{24.8}{5.3} & \underline{\res{67.8}{5.4}} & \res{27.9}{8.7} & \res{5.1}{2.1} & \underline{\res{31.2}{9.2}} & 35.2 & \underline{60.0} \\
& TRIANGLE~\cite{cicchetti2025triangle} & -  & -  & \res{68.7}{4.2} & -  & -  & \res{59.9}{3.7} & -  & -  & \res{21.9}{5.0} & -- & 50.2 \\
& \textbf{ConFu} & \cellcolor{green!15}\textbf{\res{73.8}{3.2}} & \cellcolor{green!15}\textbf{\res{33.6}{3.1}} & \cellcolor{green!15}\underline{\res{79.6}{3.6}} & \cellcolor{green!15}\textbf{\res{74.4}{2.7}} & \cellcolor{green!15}\textbf{\res{28.1}{3.2}} & \cellcolor{green!15}\textbf{\res{74.6}{4.1}} & \cellcolor{green!15}\textbf{\res{33.9}{4.2}} & \cellcolor{green!15}\textbf{\res{28.0}{2.9}} & \cellcolor{green!15}\textbf{\res{33.5}{3.2}} & \cellcolor{green!15}\textbf{45.3} & \cellcolor{green!15}\textbf{62.6} \\
\bottomrule
\end{tabular}
}
\label{tab:retrieve_multibench}
\end{table*}

\subsection{Affective Computing Benchmarks}
We next evaluate on the {MOSI}, {UR-FUNNY}, and {MUStARD} datasets, which involve text, audio, and video for multimodal affect understanding.
We assess (i) 1$\rightarrow$1 retrieval, (ii) 2$\rightarrow$1 retrieval, and (iii) multimodal classification via a linear probe.
Table~\ref{tab:retrieve_multibench} shows that ConFu consistently improves cross-modal retrieval performance. In particular, the 2$\rightarrow$1 retrieval setting surpasses 1$\rightarrow$1 retrieval on UR-FUNNY and MUStARD, suggesting that combining two input modalities provides richer and more discriminative cues. 
%These results indicate that ConFu effectively exploits complementary multimodal information rather than relying on redundant correlations, leading to stronger alignment and more stable performance across runs.
\begin{table}[t]
\centering
\caption{{Multimodal classification accuracy (\%) (mean $\pm$ std over 5 runs). Tri-CLIP refers to CLIP trained with pairwise contrastive losses. Best performance is reported in bold.}}
\resizebox{\columnwidth}{!}{
\begin{tabular}{lccc}
\toprule
\textbf{Method} & \textbf{MOSI} & \textbf{UR-FUNNY} & \textbf{MUStARD} \\
\midrule
Tri-CLIP & \sres{63.5}{2.1} & \sres{64.0}{0.9} & \sres{62.1}{3.8} \\
Symile ~\cite{saporta2024contrasting} & \textbf{\sres{67.5}{1.3}} & \sres{64.7}{1.0} & \sres{60.5}{5.0} \\
GRAM ~\cite{cicchetti2024gramian} & \sres{65.5}{2.5}   & \sres{64.8}{0.9}   & \textbf{\sres{64.7}{2.3}} \\
TRIANGLE ~\cite{cicchetti2025triangle} &  \sres{65.0}{2.8} &   \sres{64.5}{0.4} & \sres{64.6}{2.5} \\
\textbf{ConFu} & \sres{66.7}{2.1} & \textbf{\sres{64.9}{1.0}} & \sres{64.1}{2.5} \\
\bottomrule
\end{tabular}
}
\label{tab:all_classification}
\end{table}
Notably, GRAM, the only baseline supporting both retrieval configurations, struggles with 1$\rightarrow$1 alignments in general and fails for modalities M2 and M3 on MOSI and UR-FUNNY, suggesting that balancing 2$\rightarrow$1 and 1$\rightarrow$1 alignment is non-trivial.  Furthermore, methods such as TRIANGLE and Symile, which focus on trimodal alignment without explicitly modeling pairwise relations, perform consistently worse than ConFu. Overall, ConFu ranks first or second across nearly all query$\rightarrow$target retrieval settings, in both 1$\rightarrow$1 and 2$\rightarrow$1 configurations, demonstrating robust and consistent cross-modal alignment. Moreover, Table~\ref{tab:all_classification} reports linear-probe classification results, showing that ConFu attains the highest accuracy on UR-FUNNY and remains competitive on MUStARD and MOSI.
\newcommand{\ml}[1]{\large \textbf{#1}} % large, bold modality label

\subsection{Fine-Grained Bird Classification}
\label{sec_datasets:birds}
% {Finally, we assess fine-grained classification on {VB100}, {SSW60}, and {CUB200}.
% Models are pretrained on 150K audio–image–text triplets and evaluated under zero-shot and few-shot regimes.
% For videos, following~\cite{van2022exploring}, each clip is divided into eight segments and the mean frame embedding is used. Table~\ref{tab:zero_shot_results_compact} shows zero-shot results.
% ConFu achieves the highest accuracy on SSW60 (71.44\%), confirming the benefit of multimodal fusion when both modalities are informative.
% On VB100, where the audio modality is noisy (yielding only 4\% accuracy), zero-shot performance of the fused (A+V) remains close to the best visual baseline (18.15\% vs.\ 20.69\%), indicating robustness under weak modalities.
% Few-shot results presented in Fig.~\ref{fig:fewshot_linear_probing_multi_frame} demonstrate comparable gains on SSW60 and a slight performance drop on VB100, indicating that fused representations are advantageous in few-shot linear probing when both modalities provide informative signals and remain stable when they do not.}
\begin{table}[t]
\centering
\caption{
Zero-shot classification accuracy (\%) on SSW60 and VB100 datasets.  
A = Audio, V = Vision, A+V = Audio-Visual fusion. Best performance is reported in bold.
}
\resizebox{\columnwidth}{!}{%
\begin{tabular}{lccc@{\hskip 1em}ccc}
\toprule
 & \multicolumn{3}{c}{\textbf{SSW60 Acc. (\%)}} & \multicolumn{3}{c}{\textbf{VB100 Acc. (\%)}} \\
\cmidrule(lr){2-4} \cmidrule(lr){5-7}
\textbf{Method} & \textbf{A} & \textbf{V} & \textbf{A+V} & \textbf{A} & \textbf{V} & \textbf{A+V} \\
\midrule

CLIP & 29.9 & 70.1 & – & 4.2 & 20.6 & – \\
Tri-CLIP & 31.1 & 69.0 & – & 3.9 & \textbf{20.7} & – \\
Symile~\cite{saporta2024contrasting} & - & - & 60.2 & - & - & 13.4 \\
TRIANGLE ~\cite{cicchetti2025triangle} & - & - & 64.1 & - & - & 12.1 \\
GRAM ~\cite{cicchetti2024gramian} & 0.7 & 66.6 & 56.9 & 1.3 & 13.7 & 8.0 \\
\textbf{ConFu} & 30.3 & 69.4 & \textbf{71.4} & 3.4 & 19.3 & 18.1 \\
\bottomrule
\end{tabular}
}
\label{tab:zero_shot_results_compact}
\end{table}

\begin{figure}[t]
    \centering
    \begin{subfigure}{\columnwidth}
        \centering
        \includegraphics[width=0.9\linewidth]{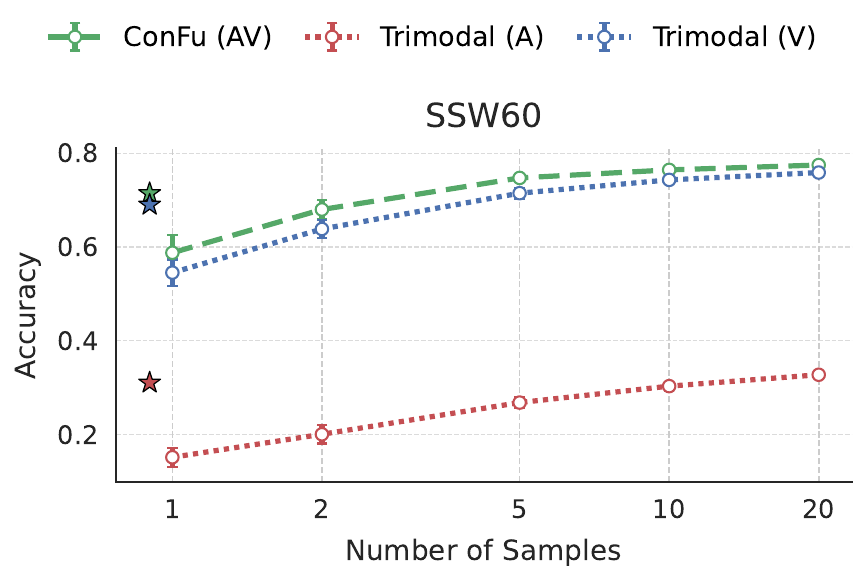}
        \vspace{2pt}
    \end{subfigure}
    
    \begin{subfigure}{\columnwidth}
        \centering
        \includegraphics[width=0.9\linewidth]{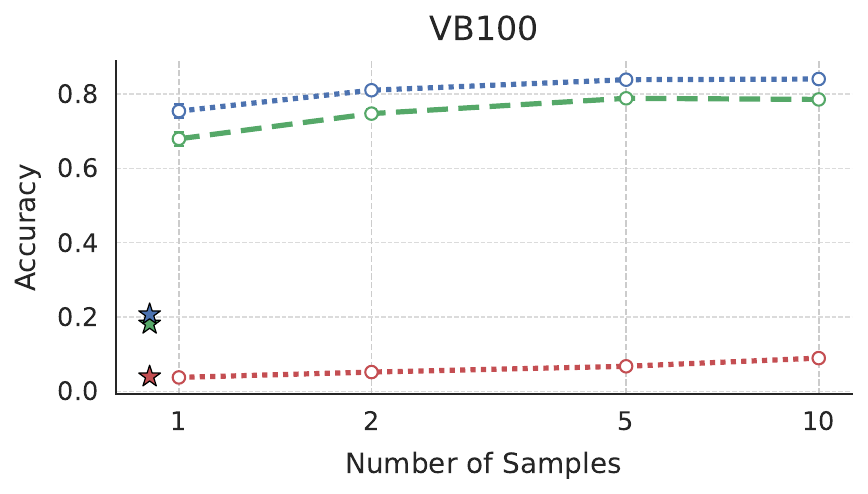}
    \end{subfigure}
    
    \caption{Few-shot linear probing results on the {SSW60} (top) and {VB100} (bottom) datasets. 
    Performance is shown as the number of labeled examples increases. 
    Zero-shot performance is indicated with a star. 
    Prediction is done in the 8-frame average embedding setting.}
    \label{fig:fewshot_linear_probing}
\end{figure}

ConFu is further evaluated on fine-grained bird classification using the {VB100} and {SSW60} datasets. Models are pretrained on Bird-MML, our curated dataset of 150K audio–image–text triplets and assessed under zero-shot and few-shot regimes. We consider two evaluation settings: (1) the multi-frame protocol from~\cite{van2022exploring}, where each video clip is divided into eight segments, the middle frame of each segment is sampled, and the mean embedding of the eight frames is used as the visual representation; and (2) a single-frame setting, providing a more constrained visual input per video. Table~\ref{tab:zero_shot_results_compact} reports zero-shot results in the multi-frame setting. ConFu achieves the highest accuracy on SSW60 (71.4\%), confirming that multimodal fusion improves fine-grained discrimination when both modalities are informative. On VB100, where the audio modality is largely uninformative ($\sim$$4\%$ unimodal accuracy), the fused model (A+V) maintains performance close to the best visual-only baseline (18.15\% vs. 20.69\%), demonstrating robustness to distracting modalities. In contrast, GRAM, Symile, and TRIANGLE exhibit a notable decline in accuracy (5–10\%) relative to ConFu.
Few-shot trends in Fig.~\ref{fig:fewshot_linear_probing} mirror these findings. Fusion offers clear advantages on SSW60 and stable performance on VB100, showing that ConFu generalizes well even when one modality is weak.

\begin{table}[t]
\centering
\caption{
Average zero-shot classification accuracy (\%) on SSW60 and VB100 datasets (mean $\pm$ std over 19 uniformly sampled frames). The full 10-second audio is used; therefore, its accuracy remains constant across samples, and no std
is reported.
A = Audio, V = Vision, A+V = Audio–Visual fusion. Best performance is reported in bold.
}
\resizebox{\columnwidth}{!}{%
\begin{tabular}{lccc@{\hskip 1em}ccc}
\toprule
 & \multicolumn{3}{c}{\textbf{SSW60 Acc. (\%)}} 
 & \multicolumn{3}{c}{\textbf{VB100 Acc. (\%)}} \\
\cmidrule(lr){2-4} \cmidrule(lr){5-7}
\textbf{Method} & \textbf{A} & \textbf{V} & \textbf{A+V} 
& \textbf{A} & \textbf{V} & \textbf{A+V} \\
\midrule
CLIP & 29.9 & \sres{60.3}{3.4} & – & 4.2 & \textbf{\sres{18.2}{0.5}} & – \\
Tri-CLIP & 31.1 & \sres{59.7}{3.2} & – & 3.9 & \sres{17.7}{0.5} & – \\
Symile~\cite{saporta2024contrasting} & - & -  & \sres{61.4}{0.2} & -& - & \sres{13.2}{0.4}\\
TRIANGLE ~\cite{cicchetti2025triangle} & -& - & \sres{58.6}{3.3} & - & - & \sres{8.7}{0.4} \\
GRAM ~\cite{cicchetti2024gramian} & 0.7 & \sres{59.0}{3.3} & \sres{54.9}{3.2} & 1.3 & \sres{12.9}{0.4} & \sres{7.6}{0.3} \\
\textbf{ConFu} & 30.3 & \sres{59.8}{3.3} & \textbf{\sres{65.5}{2.5}} 
& 3.4 & \sres{16.7}{0.5} & \sres{16.4}{0.4} \\
\bottomrule
\end{tabular}
}
\label{tab:ssw60_results_19s_compact}
\end{table}

% The CUB200 results further confirm that the learned vision encoder transfers effectively to unseen visual categories, performing on par or slightly better than the trimodal baseline despite being trained jointly with audio and text.
In the single-frame setting, designed to evaluate performance with limited visual information, we averaged classification results across 19 uniformly sampled frames to reduce bias. As provided in Table~\ref{tab:ssw60_results_19s_compact}, the fused (A+V) model exceeds the best baseline by roughly 4\% on SSW60 and complementary audio–visual cues when visual input is sparse. On VB100, performance drops slightly (by 2–3\%) due to the weaker audio modality ($\sim 4\%$ accuracy). Overall, ConFu benefits from richer multimodal supervision, maintaining strong robustness under modality degradation while preserving scalability and simplicity. Additional results on a few-shot adaptation are in the Appendix \ref{appendix:additional_results_few_shot}.

\section{Ablation}

\begin{figure}[t]
    \centering
    \includegraphics[width=0.9\linewidth]{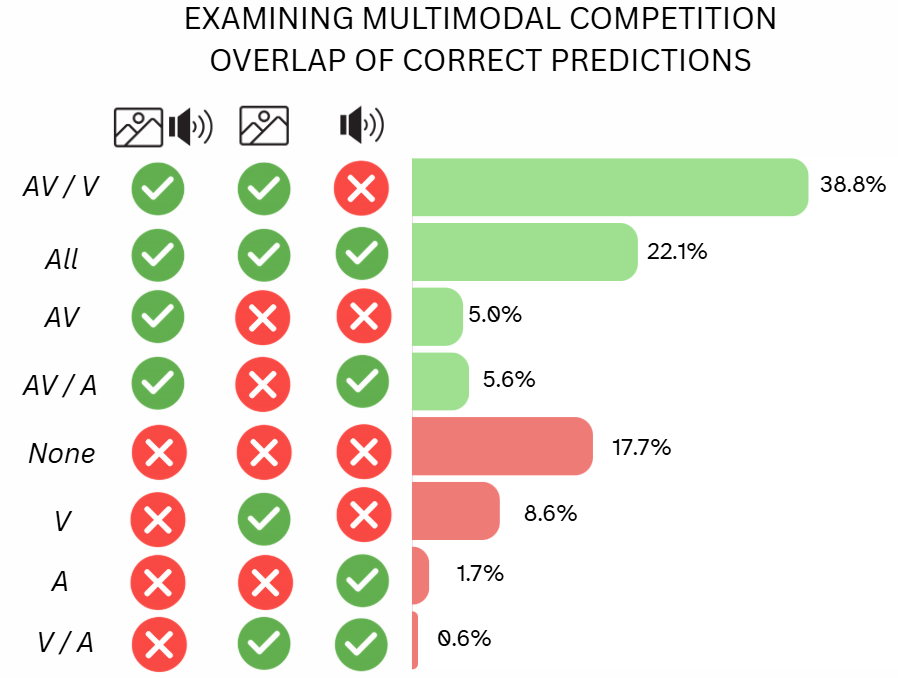}
    \caption{
    Aggregated modality overlap proportions across all classes in the SSW60 dataset.
    The chart shows the overall fraction of correctly predicted samples belonging to each overlap category:
    \textit{Audiovisual Only (AV)}, \textit{Vision Only (V)}, \textit{Audio Only (A)}, 
    \textit{Audiovisual \& Vision (AV / V)}, \textit{Audiovisual \& Audio (AV / A)}, 
    \textit{Vision \& Audio (V / A)}, \textit{All}, and \textit{None}.
    }
    \label{fig:overlap_pie}
\end{figure}

\paragraph{Modality competition.}
Although multimodal learning seeks to combine complementary cues, models often show modality dominance, where stronger signals suppress weaker ones~\cite{huang2022modality}.
To assess how our fusion is affected by this effect, we analyze in Fig.~\ref{fig:overlap_pie} the overlap of prediction errors across modalities and fusion. Results reveal visual dominance: The larger portion of correct samples (39\%) are predicted by both the fusion and vision models, while the datapoints that only audio predicts correctly or fusion-audio overlap remain small (5–6\%), and purely audio-driven successes are rare (2\%). Nevertheless, about 5\% of samples are uniquely solved by the audiovisual model, indicating that fusion captures complementary information beyond either modality alone. Finally, some classes benefit substantially from fusion (e.g., Label~31 see Appendix \ref{appendix:ablation_extra_competition}), while others remain uniformly difficult. 

\paragraph{Ablation on Noise-Induced Distribution Shift.}
To examine the sensitivity of each model to distribution shift, we introduced Gaussian noise to the image modalitiy only during testing, with a standard deviation of 0.05 corresponding to roughly 20dB SNR. This perturbation simulates a mismatch between the training and test distributions without retraining the models. As shown in Table \ref{tab:accuracy_results_noise}, {ConFu} achieves the highest accuracy (45.4\%) under these conditions. In contrast, the unimodal baseline (Tri-CLIP), exhibits a sharper performance drop, while Symile shows moderate robustness. Interestingly, TRIANGLE and GRAM perform even worse compared to Tri-CLIP under visual degradation. Overall, these results indicate that ConFu’s fusion strategy improves stability under distribution shifts affecting the image modality. Regarding more experiments in noise induced distribution shifts see Appendix \ref{appendix:ablation_extra_noise}.

\section{Discussion}
\label{sec:discussion}
% ConFu addresses a central limitation of prior multimodal frameworks: pairwise contrastive objectives capture correlations between modalities but neglect the higher-order dependencies that emerge when multiple modalities interact. By aligning fused modality pairs with a third modality, ConFu learns representations that preserve both pairwise and synergistic structure within a single embedding space.
\begin{table}[t]
\centering
\caption{
Accuracy (\%) under Gaussian noise-induced distribution shift (mean $\pm$ std over 5 runs). Noise was applied to the \textbf{Vision} modality (SNR $\sim 20dB$) at test time. A = Audio, V = Vision, A+V = Audio–Visual fusion. Best performance is reported in bold.
}
\resizebox{0.6\columnwidth}{!}{%
\begin{tabular}{lcc}
\toprule
\textbf{Method} & \textbf{Mod.} & \textbf{V noise} \\
\midrule
Tri-CLIP (V) & V & \sres{35.4}{0.4} \\
Symile~\cite{saporta2024contrasting} & A+V & \sres{40.9}{0.3} \\
GRAM~\cite{cicchetti2024gramian} & A+V & \sres{25.9}{0.1} \\
TRIANGLE~\cite{cicchetti2025triangle} & A+V & \sres{26.5}{0.2} \\
\textbf{ConFu} & A+V & \textbf{\sres{45.4}{0.1}} \\
\bottomrule
\end{tabular}
}
\label{tab:accuracy_results_noise}
\end{table}

Empirical results across synthetic and real-world datasets demonstrate that ConFu effectively captures complementary cross-modal structure. On the synthetic XOR benchmark, ConFu successfully models the underlying multimodal dependencies that pairwise and geometric baselines miss even with higher-dimensional representations (Appendix \ref{appendix:ablation_xor}). In AV-MNIST, ConFu dominates unimodal approaches in zero-shot classification, indicating successful utilization of both modalities. The 2$\rightarrow$1 retrieval evaluations on UR-FUNNY, and MUStARD further confirm that combining two query modalities provides richer semantic grounding than either considered in isolation. An interesting observation in the 2$\rightarrow$1 case on the MOSI dataset is that fusion slightly underperforms compared to the pairwise baseline. Remarkably, this issue is substantially alleviated by introducing partial modality masking (Appendix~\ref{appendix:ablation_extra_mask}), supporting the hypothesis that suboptimal performance stems from shortcut learning in cases where redundant information dominates.

In fine-grained bird classification, ConFu maintains or slightly exceeds unimodal baselines. On SSW60, where both modalities contribute meaningfully, fusion achieves the best accuracy, on VB100, where acoustic information is limited, performance remains close to the visual baseline. Moreover, modality overlap analyses indicates that a subset of samples is correctly resolved only when both modalities are utilized, reflecting the model’s ability to integrate complementary evidence.  Finally, under visual perturbations, ConFu degrades more gracefully, outperforming all baselines by a clear margin. SYMILE ranks second, remaining above unimodal performance, while TRIANGLE and GRAM drop below the single-modality baseline. 

\textbf{Limitations and Future Work.} ConFu is well-positioned to scale to any number of modalities, offering flexibility for diverse multimodal tasks. However, as the number of modalities increases, computational demands may also rise. Depending on the task, selectively pruning certain alignment terms, shown in Appendix~\ref{app:generalM}, can help maintain efficiency. For example, omitting many-to-many alignments, when assuming that only single-modality queries or targets will be needed, opens a promising avenue for future adaptation and research. 

ConFu currently relies on fully aligned modalities during training, and exploring ways to relax or bypass this requirement remains crucial for many applications. In particular, enabling ConFu to handle partially missing modalities, such as training with incomplete triplets or drawing on ideas related to pseudopair construction  \cite{wang2024omnibind, zhang2024extending, wang2023connecting}, could potentially make the framework more robust and broadly applicable across real-world multimodal settings. Further prospective directions could include extending ConFu to incorporate factorized representations inspired by recent factorization-based and redundancy-reduction approaches~\cite{liang2023factorized, wang2024decoupling, dufumier2024align}, developing adaptive mechanisms to regulate modality competition~\cite{kontrasbalancing, MLB, OGMGE, ReconBoost, wei2024diagnosing}, and investigating how ConFu can enhance robustness in multimodal retrieval, particularly when one or more modalities are noisy, incomplete, or severely corrupted.

\section{Conclusion}
\label{sec:conclusion_final}
We presented ConFu, a unified contrastive framework that jointly optimizes pairwise and higher-order alignments to capture pairwise and synergistic relationships across modalities. The method supports both 1→1 and 2→1 retrieval, improving multimodal alignment and robustness without modifying encoder architectures. Overall, ConFu demonstrates strong adaptability across diverse datasets, domains, and challenging conditions such as noise-induced distribution shifts, modality degradation, and missing modalities, while introducing only minimal overhead through lightweight MLPs atop the encoders, establishing it as a robust and scalable solution for multimodal alignment in real-world applications. The accompanying Bird-MML dataset establishes a controlled setup for studying multimodal complementarity under realistic audio–visual conditions.

\clearpage

\section{Acknowledgments}
This work was supported by the HARSH project (project no. Y$\Pi$3TA - 0560901) within the framework
of the National Recovery and Resilience Plan--Greece 2.0--with funding from the European Union--NextGenerationEU; the Hellenic Foundation for Research and
Innovation (H.F.R.I.) under the “3rd Call for H.F.R.I.’s Research
Projects to Support Faculty Members \& Researchers” (H.F.R.I. Project
Number: 26302); by the TITAN ERA Chair project (contract no.
101086741) within the Horizon Europe Framework Program of the European
Commission; the METHUSALEM program (METH/26/003, “Methusalem-BioMedAI”); the Flemish Government AI Research Program (Leuven.AI – KU Leuven); the FWO project “Task- and Device-Agnostic Artificial Intelligence for EEG Analysis” (G046925N); and the Horizon Europe project “AI-PROGNOSIS” (Grant Agreement No. 101080581).

% This work was supported by the Greek Ministry of Education, Religious Affairs and Sports through the HARSH project (project no. Y$\Pi$3TA - 0560901), within the framework
% of the National Recovery and Resilience Plan--Greece 2.0--with funding from the European Union--NextGenerationEU. 

% This work has been supported by the TITAN ERA Chair project (contract no. 101086741) within the Horizon Europe Framework Program of the European Commission 

{
    \small
    \bibliographystyle{ieeenat_fullname}
    \bibliography{main}
}

\clearpage
\setcounter{page}{1}
\maketitlesupplementary

\appendix

\section*{Supplementary Material}

This supplementary document provides additional details and results that complement the main paper.
It is organized as follows:

\begin{itemize}
    \item \textbf{Experimental Details.} We present comprehensive descriptions of model configurations, hyperparameters, training procedures, and evaluation protocols to ensure reproducibility.
    \item \textbf{Additional Ablation Studies.} We include supplementary ablation experiments that highlight additional aspects of this work, examining how different model components and design choices influence overall performance.
    \item \textbf{Additional Results.} This section includes further quantitative results that extend and support the findings presented in the main paper.
    \item \textbf{Datasets.} Additional information regarding the datasets used and our new dataset Bird-MML.
\end{itemize}

\section{Experimental Details}
\label{appendix:experimental_details}

\subsection{Experiments on Bird-MML}

\paragraph{Dataset preprocessing}
For the audio modality,
Each audio clip was truncated or zero-padded to a duration of 10 seconds and resampled to 22.05~kHz. Mel spectrograms were computed using a window size of 1024 and a hop length of 512, resulting in 128 Mel frequency bands. Each spectrogram was treated as a single-channel image to ensure compatibility with a ResNet-50 backbone. 

For the image modality,
Images were resized to $224 \times 224$ pixels and normalized. No data augmentation was applied.

\paragraph{Model Architecture.}
The multimodal architecture employed is composed of three modality-specific encoders:
\begin{itemize}
    \item \textbf{Text encoder:} BERT transformer.
    \item \textbf{Image encoder:} ResNet-50.
    \item \textbf{Audio encoder:} ResNet-50 applied to Mel-spectrogram inputs.
\end{itemize}

Each encoder outputs a 512-dimensional embedding. A two-layer multilayer perceptron (MLP) was used for multimodal feature fusion. All models and baselines were trained using identical hyperparameters, without any additional hyperparameter search or tuning.

Training used the AdamW optimizer with a learning rate of $1\times10^{-4}$ and a weight decay of $1\times10^{-4}$. A batch size of 64 was used throughout the training process. The learning rate followed a cosine annealing schedule with a minimum of $1\times10^{-6}$. The fusion loss coefficient was set to $\lambda = 0.5$.

\subsection{MOSI, MUStARD, and UR-FUNNY}

For these datasets, all models were trained for 100 epochs, producing 256-dimensional representations. We used a batch size of 128 and a learning rate of $5\times10^{-5}$. Following prior work \cite{liang2023factorized}, each modality encoder was implemented as a 5-layer Transformer with 5 attention heads.

Fusion was achieved through a 2-layer MLP that combined unimodal embeddings into a single fused representation. Dataset splits followed the official \texttt{MultiBench} configuration for training, validation, and testing.

\subsection{AVMNIST}

For the text modality, we used a BERT transformer. Image, spectrogram, and satellite-view inputs were encoded using ResNet-18 backbones initialized from ImageNet and fully trained. Spectrograms provided by the \texttt{MultiBench} dataloaders were used without modification.

Embeddings had a 256-dimensional size. Dataset splits again followed the \texttt{MultiBench} protocol.

\paragraph{Training Setup.}
All models were trained from scratch for 30 epochs using the AdamW optimizer with a learning rate of $1\times10^{-4}$ and a batch size of 256. The best model was selected based on validation loss. Each experiment was repeated with five random seeds, and we report the mean and standard deviation across runs. For our model, MLP-based fusion was used with a balancing parameter $\lambda = 0.5$ between pairwise and fusion losses.

\subsection{XOR Experiment}

For the XOR experiment, we mapped all modalities to embeddings of size 128. The dataset consisted of 10,000 training samples and 5,000 test samples. Models were trained for 50 epochs with a batch size of 512 and a learning rate of $1\times10^{-4}$. 

Each encoder was implemented as a 2-layer MLP, and we used a balanced fusion coefficient of $\lambda = 0.5$.

% \clearpage

\section{Additional Ablation Studies}

\label{appendix:ablation_extra}

\subsection{Effect of the $\lambda$ Parameter}
\label{appendix:ablation_lambda}

As shown in Table~\ref{tab:lambda_ablation_pct}, performance is not overly sensitive to $\lambda$. Accuracy remains stable while recall varies smoothly, indicating a controlled trade-off with minimal tuning required. Identifying a principled way to set or eliminate $\lambda$ is an important direction. 

\begin{table}[h]
\caption{{Effect of $\lambda$ across datasets (mean $\pm$ std). Accuracy (\%) remains stable while $\lambda$ trades off recall. Best values in bold.
}}
\centering
% \small
\footnotesize
\setlength{\tabcolsep}{4pt}
\begin{tabular}{lcccc}
\toprule
Dataset & $\lambda$ & Acc. (\%) & $\mathrm{R}_{\text{mean}@10}^{1\to1}$ & $\mathrm{R}_{\text{mean}@10}^{2\to1}$ \\
\midrule
\multirow{3}{*}{MOSI} 
 & 0.2 & \textbf{67.7} $\pm$ 1.7 & 19.5 $\pm$ 1.0 & \textbf{20.1} $\pm$ 1.9 \\
 & 0.5 & 66.7 $\pm$ 2.1 & \textbf{19.5} $\pm$ 1.5 & 19.6 $\pm$ 2.0 \\
 & 0.8 & 65.8 $\pm$ 2.0 & 13.5 $\pm$ 2.2 & 15.6 $\pm$ 1.6 \\
\midrule
\multirow{3}{*}{UR-Funny} 
 & 0.2 & 64.2 $\pm$ 0.7 & \textbf{8.0} $\pm$ 0.4 & \textbf{14.0} $\pm$ 0.9 \\
 & 0.5 & \textbf{64.9} $\pm$ 1.0 & 7.4 $\pm$ 0.2 & 13.6 $\pm$ 0.6 \\
 & 0.8 & 64.6 $\pm$ 0.7 & 6.9 $\pm$ 0.3 & 13.2 $\pm$ 0.7 \\
\midrule
\multirow{3}{*}{MUSTARD} 
 & 0.2 & 59.7 $\pm$ 3.2 & 42.1 $\pm$ 2.2 & 57.1 $\pm$ 2.5 \\
 & 0.5 & \textbf{64.1} $\pm$ 2.5 & \textbf{45.3} $\pm$ 2.6 & \textbf{62.6} $\pm$ 3.1 \\
 & 0.8 & 61.1 $\pm$ 3.4 & 41.0 $\pm$ 1.0 & 59.7 $\pm$ 1.1 \\
\bottomrule
\end{tabular}
\label{tab:lambda_ablation_pct}
\end{table}

Regarding optimal $\lambda$ values the results in Table~\ref{tab:recall10_results} highlight the influence of the $\lambda$ parameter, which balances the contribution of modality-specific and shared representations.
These $\lambda$ values were selected from the Pareto front obtained by jointly optimizing the two retrieval objectives - mean Recall@10 for 1$\rightarrow$1 and 2$\rightarrow$1 tasks.
We observe that moderate $\lambda$ values (0.1--0.5) yield the best trade-off between the two objectives, and that the optimal $\lambda$ varies across datasets, indicating optimal $\lambda$ depends on data distribution.

\begin{table}[h!]
\centering
\caption{Mean Recall@10 results for 1$\to$1 and 2$\to$1 tasks across datasets.}
\resizebox{\columnwidth}{!}{
\begin{tabular}{lccc}
\hline
\textbf{Dataset} & \textbf{$\lambda$} & \textbf{Mean R@10 (1$\to$1)} & \textbf{Mean R@10 (2$\to$1)} \\
\hline
MOSI     & 0.1 & 21.21 & 22.12 \\
UR-FUNNY & 0.4 &  8.06 & 14.69 \\
MUSTARD  & 0.5 & 45.29 & 62.56 \\
\hline
\end{tabular}
}
\label{tab:recall10_results}
\end{table}

\subsection{Embedding Dimensionality Analysis on the XOR Task}

\label{appendix:ablation_xor}

\begin{figure}[h]
    \centering
    \includegraphics[width=0.95\linewidth]{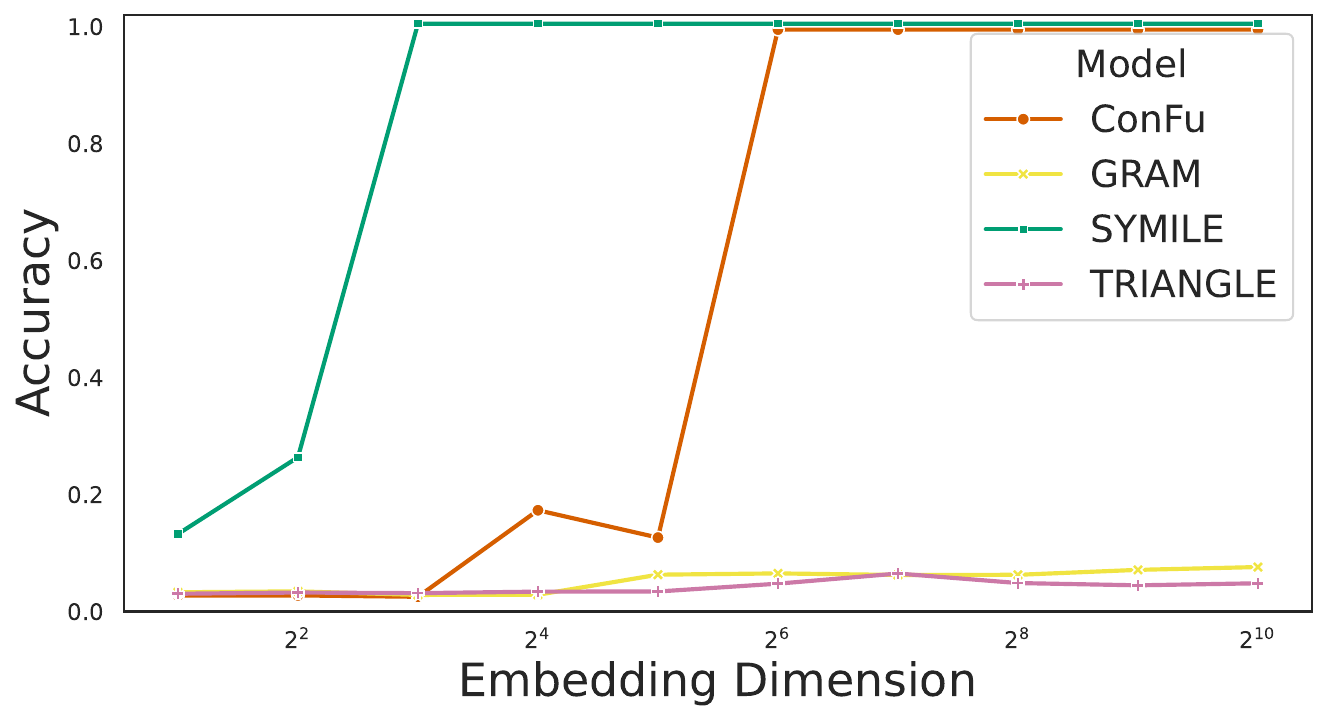}
    \caption{
        Accuracy as a function of embedding dimension for $\hat{p} = 1$.
        SYMILE achieves perfect accuracy from dimension 8 onwards, 
        while \textsc{ConFu} requires a dimensionality of 64 to converge.
        \textsc{TRIANGLE} and \textsc{GRAM} fail to solve the XOR task even at 1024 dimensions.
    }
    \label{fig:xor_embedding_dim}
\end{figure}

In this experiment, we evaluate how the embedding dimensionality affects each method’s ability to solve the XOR problem for the case of $\hat{p} = 1$. 
Figure~\ref{fig:xor_embedding_dim} illustrates the mean accuracy as a function of the embedding dimension. As provided, 
% \textbf{Results.}
SYMILE successfully learns to solve the XOR task even with a minimal embedding dimensionality of 8, achieving perfect accuracy thereafter. 
Our proposed method (\textsc{ConFu}) also manages to solve XOR but requires a larger embedding dimensionality of 64 to reach convergence. 
In contrast, both \textsc{TRIANGLE} and \textsc{GRAM} fail to solve the task even when the embedding dimensionality is increased up to 1024, 
implying that the bottleneck lies in their loss formulations rather than in representational capacity.

\subsection{Tackling modality shortcuts}

\label{appendix:ablation_extra_mask}

\begin{figure*}[h!]
\centering
% -------- (a) Accuracy --------
\begin{subfigure}[h]{0.32\textwidth}
    \centering
    \includegraphics[width=\linewidth]{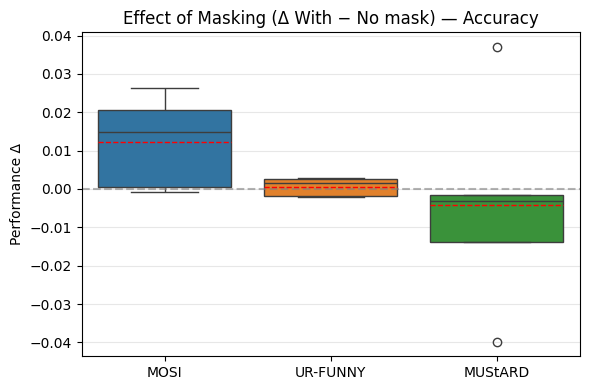}
    \caption{Performance difference ($\Delta$ With - No mask) for the \textbf{Accuracy} task. Each box shows variability across $\lambda$ values for different datasets.}
    \label{fig:mask_effect_accuracy}
\end{subfigure}
\hfill
% -------- (b) 1→1 Retrieval --------
\begin{subfigure}[h]{0.32\textwidth}
    \centering
    \includegraphics[width=\linewidth]{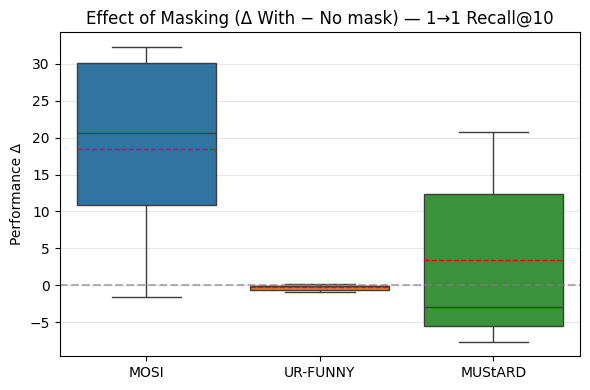}
    \caption{Performance difference for \textbf{1$\rightarrow$1 retrieval (Recall@10)}. Positive values indicate that masking improves unimodal retrieval.}
    \label{fig:mask_effect_1to1}
\end{subfigure}
\hfill
% -------- (c) 2→1 Retrieval --------
\begin{subfigure}[h]{0.32\textwidth}
    \centering
    \includegraphics[width=\linewidth]{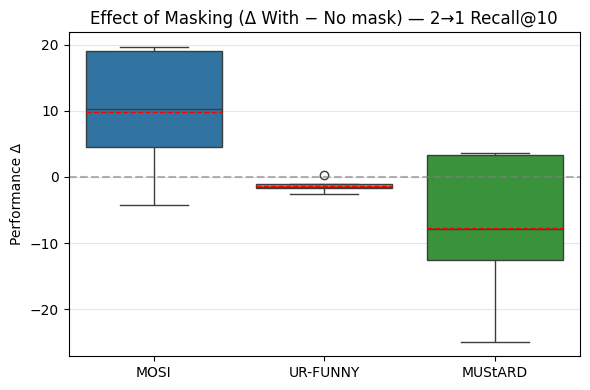}
    \caption{Performance difference for \textbf{2$\rightarrow$1 retrieval (Recall@10)}. Each box aggregates runs with different $\lambda$ values and mask ratios.}
    \label{fig:mask_effect_2to1}
\end{subfigure}

\caption{
Effect of the masking ratio on multimodal performance for the MLP fusion model.
Each subplot reports the performance change ($\Delta =$ With mask $-$ No mask) across datasets and tasks, computed over multiple $\lambda$ and mask ratio values.
These results illustrate how both hyperparameters jointly influence model behavior for classification and retrieval tasks.
}
\label{fig:mask_effect_all}
\end{figure*}

Motivated by the hypothesis that the fusion network may exploit shortcut strategies, we investigated the effect of applying dropout at the feature level, specifically, before concatenation and subsequent processing by the fusion MLP.
A potential shortcut can arise when the pairwise alignment losses between modalities become sufficiently low, allowing the fusion module to rely predominantly on one modality while suppressing the contribution of the other.
In such cases, the network may learn to transmit information through the most predictive modality instead of developing genuinely integrated multimodal representations.
Introducing feature-level masking may counteract this behavior by preventing the model from depending exclusively on any single modality and encouraging more balanced multimodal learning.

To examine this effect, we compared model performance with and without masking for multiple $\lambda$ values and mask ratios.
For each dataset and task, we computed the performance difference $\Delta = (\text{With mask}) - (\text{No mask})$ across runs and visualized these distributions using boxplots.

As shown in Figure~\ref{fig:mask_effect_all}, the MOSI dataset consistently benefits from masking across all tasks, classification as well as 1$\rightarrow$1 and 2$\rightarrow$1 retrieval, indicating that feature-level dropout can improve representation robustness and reduce reliance on modality-specific cues.
In contrast, the UR-FUNNY dataset exhibits minimal changes, suggesting that masking has limited impact on its multimodal dynamics, while MUStARD experiences performance degradation, likely due to its stronger dependence on precise cross-modal alignment. Moreover, as shown in Table~\ref{tab:masking_mosi}, the use of masking in ConFu yields significant performance gains across all query$\rightarrow$target retrieval directions on the MOSI dataset.
Overall, these findings suggest that the impact of masking is dataset-dependent: it tends to improve performance when redundant cross-modal information could otherwise enable shortcut learning, but may hinder it when successful alignment relies on using the full representational capacity of each modality.

\begin{table}[h!]
\caption{Recall@10 (\%) on MOSI with and without masking (mask ratio = 0.3, $\lambda = 0.5$). $\Delta$ indicates the absolute improvement.}
\resizebox{\columnwidth}{!}{
\begin{tabular}{l l | c c c}
\toprule
\textbf{Target} & \textbf{Query(s)} & \textbf{Without Mask} & \textbf{With Mask} & \textbf{$\Delta$} \\
\midrule
\multirow{3}{*}{\ml{M1}} 
 & \ml{M2}  & \res{21.02}{1.65} & \textbf{\res{25.10}{3.51}} & +4.08 \\
 & \ml{M3}  & \res{16.41}{2.32} & \textbf{\res{20.41}{1.51}} & +4.00 \\
 & \ml{M23} & \res{16.73}{1.84} & \textbf{\res{22.24}{3.16}} & +5.51 \\
\midrule
\multirow{3}{*}{\ml{M2}} 
 & \ml{M1}  & \res{19.18}{1.41} & \textbf{\res{25.83}{2.98}} & +6.65 \\
 & \ml{M3}  & \res{21.02}{2.40} & \textbf{\res{22.77}{0.96}} & +1.75 \\
 & \ml{M13} & \res{21.63}{1.89} & \textbf{\res{24.58}{2.33}} & +2.95 \\
\midrule
\multirow{3}{*}{\ml{M3}} 
 & \ml{M1}  & \res{16.06}{1.74} & \textbf{\res{23.35}{1.61}} & +7.29 \\
 & \ml{M2}  & \res{23.47}{2.69} & \textbf{\res{24.40}{1.91}} & +0.93 \\
 & \ml{M12} & \res{20.50}{2.87} & \textbf{\res{24.34}{3.02}} & +3.84 \\
\bottomrule
\end{tabular}
}
\label{tab:masking_mosi}
\end{table}

\subsection{Noise experiments}
\label{appendix:ablation_extra_noise}
% \label{appendix:noise_ablation}

We investigate the robustness of ConFu under controlled noise-induced distribution shifts applied to either the image or audio modality. Gaussian noise was added at varying severities with standard deviations of 0.05, 0.1, and 0.15 for images, and 0.1, 0.2, and 0.3 for audio. The corresponding SNR values are estimated from the data.

The results in Table \ref{tab:accuracy_results_noise_multiple} show that ConFu exhibits improved robustness to noise compared to competing baselines. Notably, GRAM and TRIANGLE appear largely unaffected when noise is added to audio. However, this behavior is likely attributable to their effective disregard of the audio modality. In contrast, both methods degrade substantially when noise is introduced to the image modality, performing even worse than unimodal TriCLIP.

Overall, ConFu consistently outperforms all baselines across noise levels, including unimodal variants that do not suffer degradation. Its performance trails unimodal TriCLIP-audio slightly only in the extreme case where the visual modality becomes heavily corrupted. A key observation is that Symile, GRAM, and TRIANGLE deteriorate markedly under noise, whereas ConFu incurs only minimal performance loss, particularly when noise is added to audio, where the accuracy drop is negligible. For comparison, TriCLIP experiences roughly a 5\% accuracy decline at an audio SNR of 10 dB, while ConFu drops by only $\sim$0.2\%.

\begin{table}[t]
\centering
\caption{
Accuracy (\%) under modality-specific noise-induced distribution shift. Noise was applied to individual modalities at test time with varying SNR levels. A = Audio, V = Vision, A+V = Audio–Visual fusion. Best performance for each degradation type and level is reported in bold.
}
\resizebox{\columnwidth}{!}{%
\begin{tabular}{lcccccc}
\toprule
 & \multicolumn{2}{c}{\textbf{10dB SNR}} & \multicolumn{2}{c}{\textbf{15dB SNR}} & \multicolumn{2}{c}{\textbf{20dB SNR}} \\
\cmidrule(lr){2-3} \cmidrule(lr){4-5} \cmidrule(lr){6-7}
\textbf{Method} & \textbf{A deg.} & \textbf{V deg.} & \textbf{A deg.} & \textbf{V deg.} & \textbf{A deg.} & \textbf{V deg.} \\
\midrule
Tri-CLIP (V) & 69.0 & 5.5 & 69.0 & 13.2 & 69.0 & 35.4 \\
Tri-CLIP (A) & 26.0 & \textbf{31.1} & 29.1 & 31.1 & 30.3 & 31.1 \\
Symile~\cite{saporta2024contrasting} & 58.4 & 21.2 & 59.5 & 27.9 & 60.0 & 40.9 \\
GRAM~\cite{cicchetti2024gramian} & 58.9 & 4.0 & 58.3 & 8.94 & 58.1  & 25.9 \\
TRIANGLE~\cite{cicchetti2025triangle} & 63.9 & 3.4 & 64.0 & 7.24 & 64.0 & 26.5 \\
\textbf{ConFu} & \textbf{71.2} & 30.2 & \textbf{71.4} & \textbf{33.1} & \textbf{71.5} & \textbf{45.4} \\
\bottomrule
\end{tabular}%
}
\label{tab:accuracy_results_noise_multiple}
\end{table}

% \begin{figure}[t]
%     \centering
%     \includegraphics[width=\columnwidth]{figs/supplementary_noise_images/noise-normal.png}
%     \caption{
%     Noise on the image on the left clean image on the right.
%     }
%     \label{fig:robustness-noise}
% \end{figure}

\subsection{Discussion on Multimodal Competition}
\label{appendix:ablation_extra_competition}
% \label{app:label_competition}

\begin{figure*}[t]
    \centering
    \includegraphics[width=0.65\linewidth]{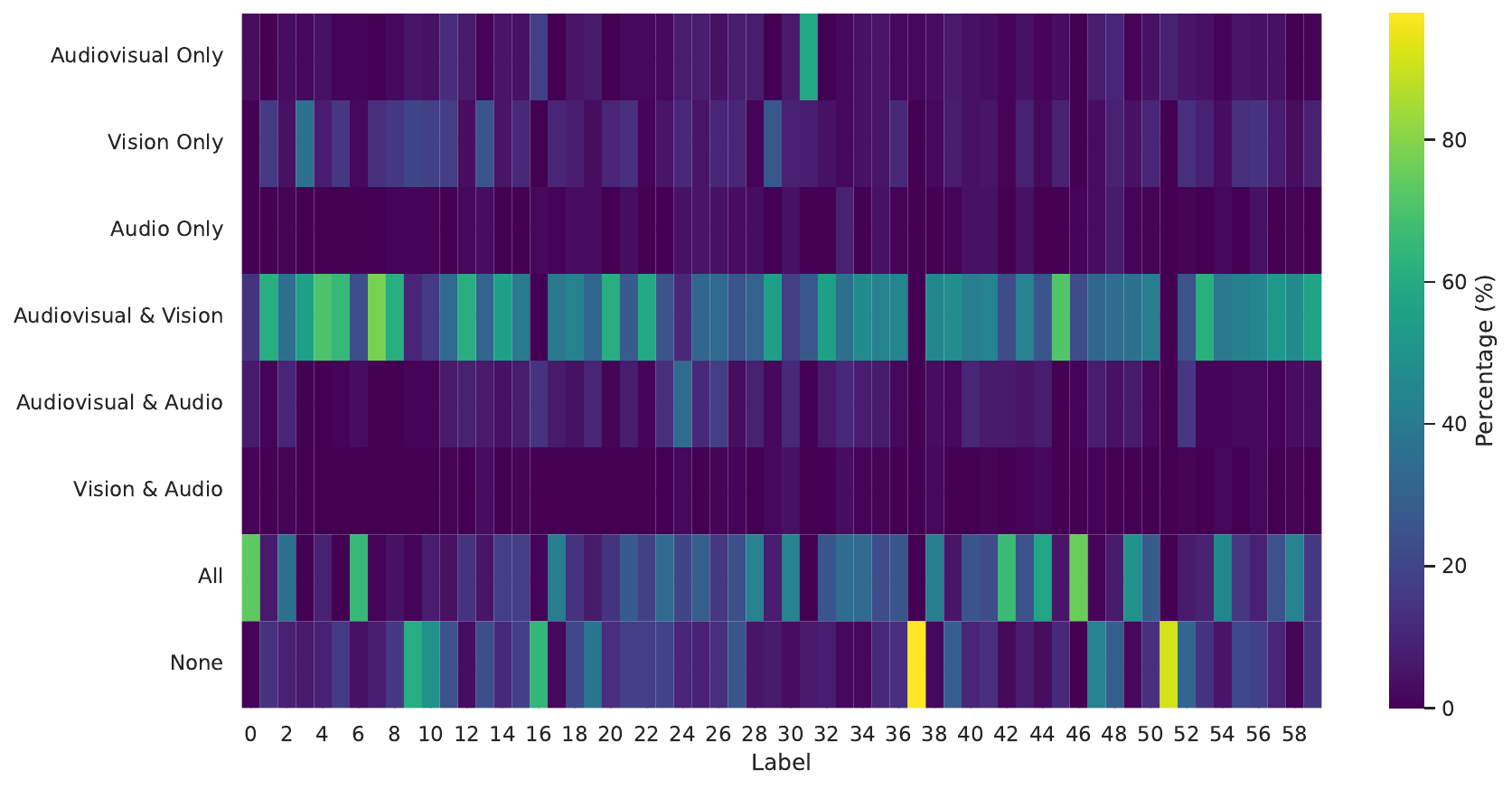}
    \caption{
    Per-class modality overlap analysis on the SSW60 dataset in the zero-shot classification setting.
    Each cell shows the percentage of samples within a class that fall into a given overlap category:
    \textit{Audiovisual Only}, \textit{Vision Only}, \textit{Audio Only}, 
    \textit{Audiovisual \& Vision}, \textit{Audiovisual \& Audio}, 
    \textit{Vision \& Audio}, \textit{All}, or \textit{None}.
    Higher values indicate a larger fraction of samples for which that modality (or combination) produces correct predictions.
    This visualization highlights class-specific complementarity and competition between modalities.
    }
    \label{fig:overlap_heatmap}
\end{figure*}

In Fig.~\ref{fig:overlap_heatmap}, we present a detailed per-class modality overlap analysis for the SSW60 dataset in the zero-shot classification setting. This experiment visualizes, in the form of a heatmap, the distribution of overlap categories for each label separately. Specifically, each cell indicates the percentage of samples belonging to a given class that fall into one of the following overlap categories: \emph{Audiovisual Only}, \emph{Vision Only}, \emph{Audio Only}, \emph{Audiovisual \& Vision}, \emph{Audiovisual \& Audio}, \emph{Vision \& Audio}, \emph{All}, or \emph{None}. These categories capture whether a sample is correctly classified by only one modality, by a specific combination of modalities, by all modalities, or not correctly classified at all.

Higher values in a cell correspond to a larger proportion of samples for which that modality (or modalities) yields a correct prediction. This visualization allows us to inspect class-specific modality behavior, highlighting both complementary and competing contributions across modalities.

We observe that label~31 is exclusively correctly classified by the \emph{Audiovisual Only} modality, suggesting that neither vision nor audio alone provides sufficient information to recognize this class. This reinforces the need for complementary cross-modal information in order to correctly classify certain bird species.

% \clearpage

\section{Additional Results}
\label{appendix:additional_results}

\subsection{Few shot adaptation on SSW60, VB100, CUB200 complete results}
\label{appendix:additional_results_few_shot}

In this section, we present the complete results of our few-shot adaptation experiments on the SSW60, VB100, and CUB200~\cite{wah2011caltech} datasets. The following figures include all baseline methods for comparison and illustrate adaptation performance across varying number of shot settings. These results provide a comprehensive view of how each method generalizes to limited-data scenarios and adapts to novel classes.

\paragraph{SSW60.}
In the multi-frame setting, where mean embeddings over 8 video frames are used, our fusion-based method achieves the best performance. It is important to note that competing approaches do not support multimodal representations, and therefore only unimodal results are available for comparison. In the vision modality, most methods perform similarly, with Symile showing a slight drop. In contrast, in the audio modality, TRIANGLE and GRAM underperform significantly, indicating that their generated audio representations are not informative. In the single-frame setting, \textbf{ConFu}'s audiovisual representations outperform all unimodal variants by a substantial margin.

\paragraph{VB100.}
In VB100, audiovisual fusion performs worse than vision-only models. This aligns with our earlier discussion: the audio modality is largely uninformative and even distracting for this dataset, as reflected by the low performance of audio-only baselines.

\paragraph{CUB200.}
For CUB200, we evaluate on classes completely unseen during pretraining. Symile achieves the strongest performance, while our method (\textbf{ConFu}) ranks second.

\begin{figure*}[h]
    \centering
    \includegraphics[width=0.9\textwidth]{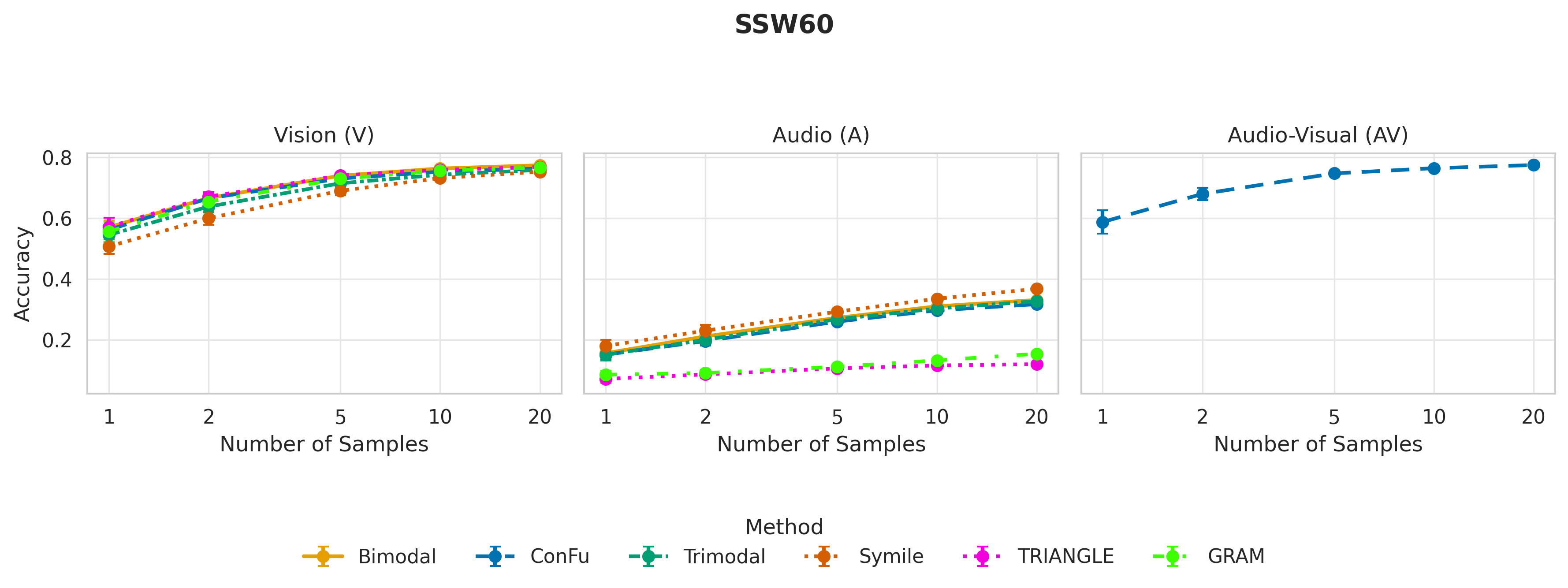}
    \caption{Few-shot linear probing results for SSW60~\cite{van2022exploring}. Performance is shown as the number of labeled examples increases. Prediction is done in the 8-frame sampling setting (multi-frame).}
    \label{fig:suppl_fewshot_ssw60_multiframe}
\end{figure*}

\begin{figure*}[h]
    \centering
    \includegraphics[width=0.9\textwidth]{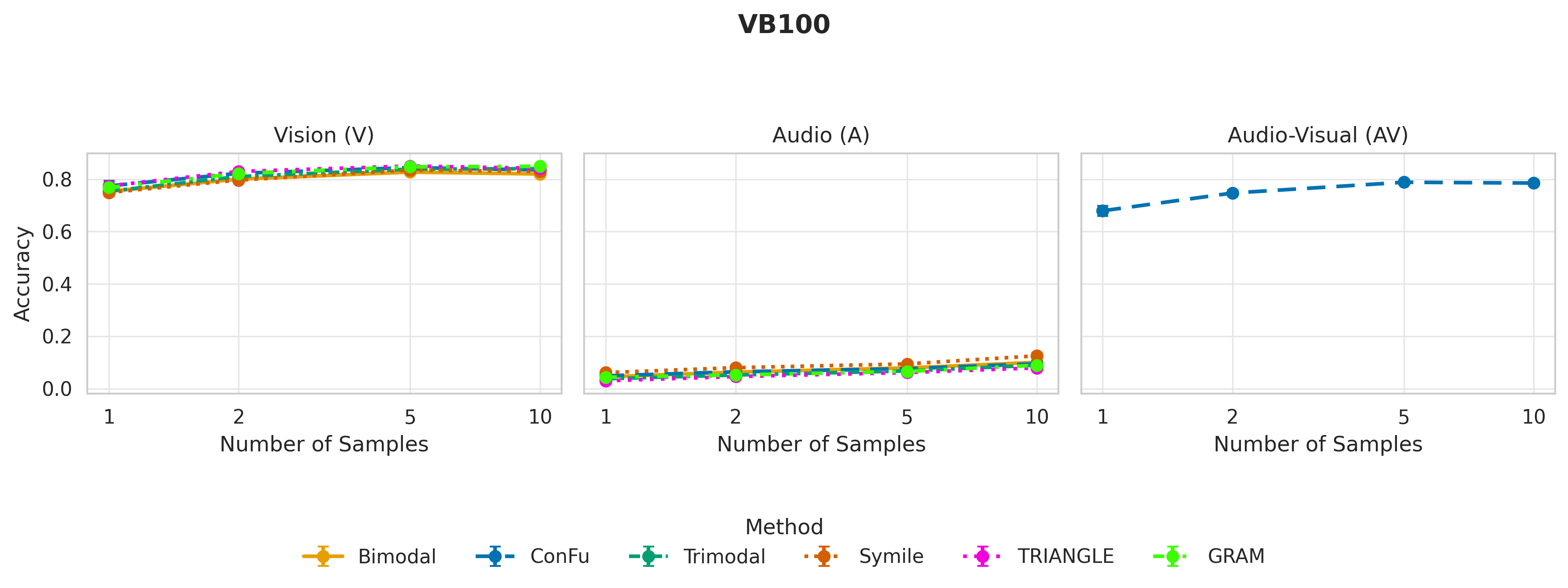}
    \caption{Few-shot linear probing results for VB100~\cite{ge2016exploiting}. Performance is shown as the number of labeled examples increases. Prediction is done in the 8-frame sampling setting (multi-frame).}
    \label{fig:suppl_fewshot_vb100_multiframe}
\end{figure*}

\begin{figure*}[h]
    \centering
    \includegraphics[width=0.9\textwidth]{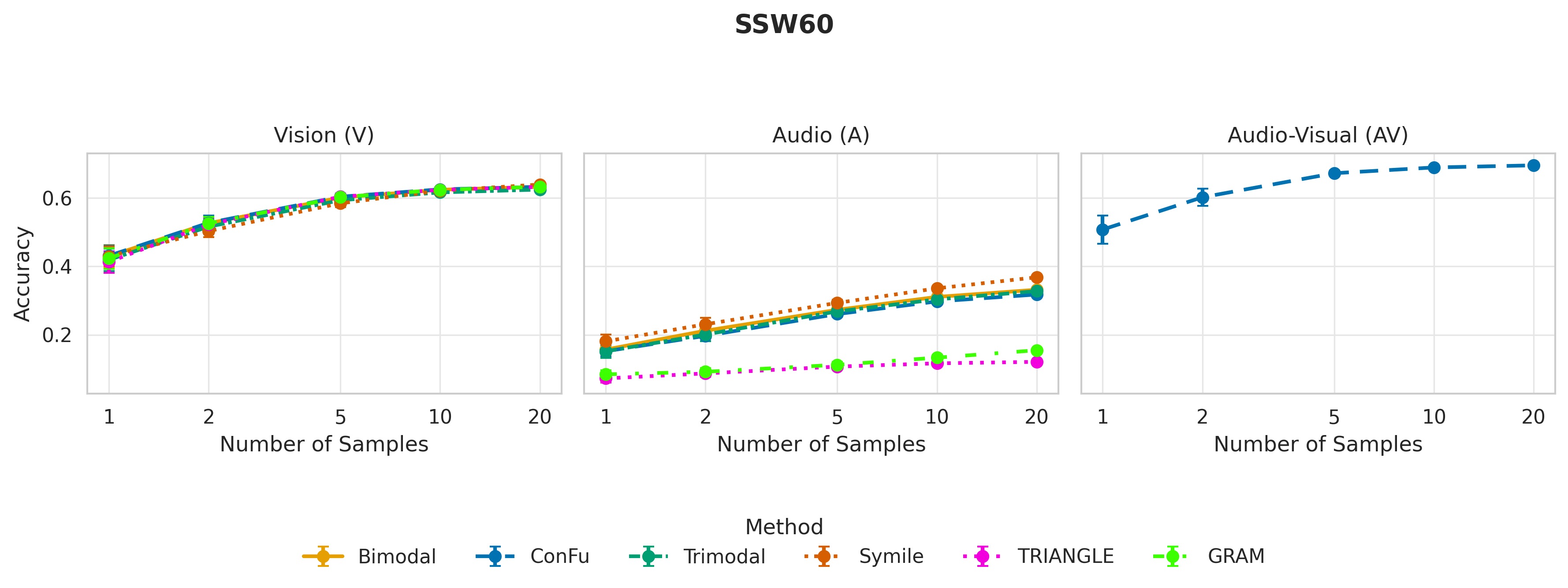}
    \caption{Few-shot linear probing results for SSW60~\cite{van2022exploring}. Performance is shown as the number of labeled examples increases.Prediction is done in the single frame setting.}
    \label{fig:suppl_fewshot_ssw60_singleframe}
\end{figure*}

\begin{figure*}[h]
    \centering
    \includegraphics[width=0.9\textwidth]{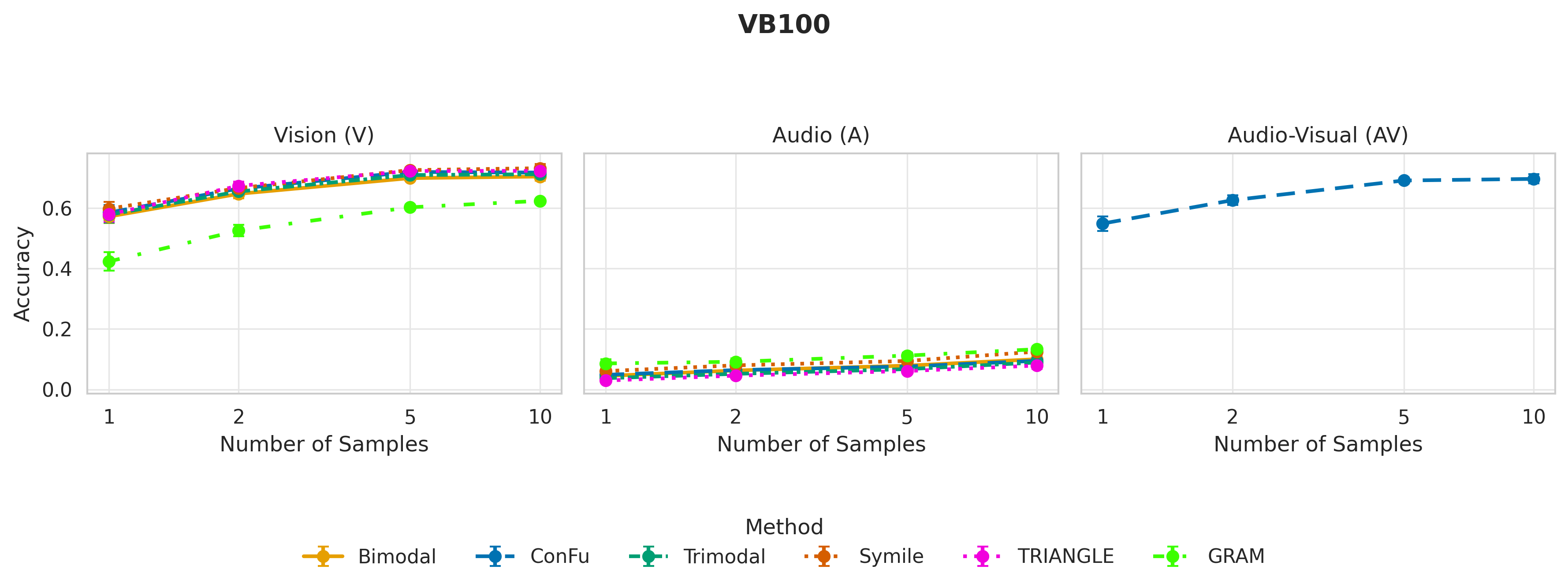}
    \caption{Few-shot linear probing results for VB100~\cite{ge2016exploiting}. Performance is shown as the number of labeled examples increases.Prediction is done in the single frame setting.}
    \label{fig:suppl_fewshot_vb100_singleframe}
\end{figure*}

\begin{figure}[h]
    \centering
    \includegraphics[width=\columnwidth]{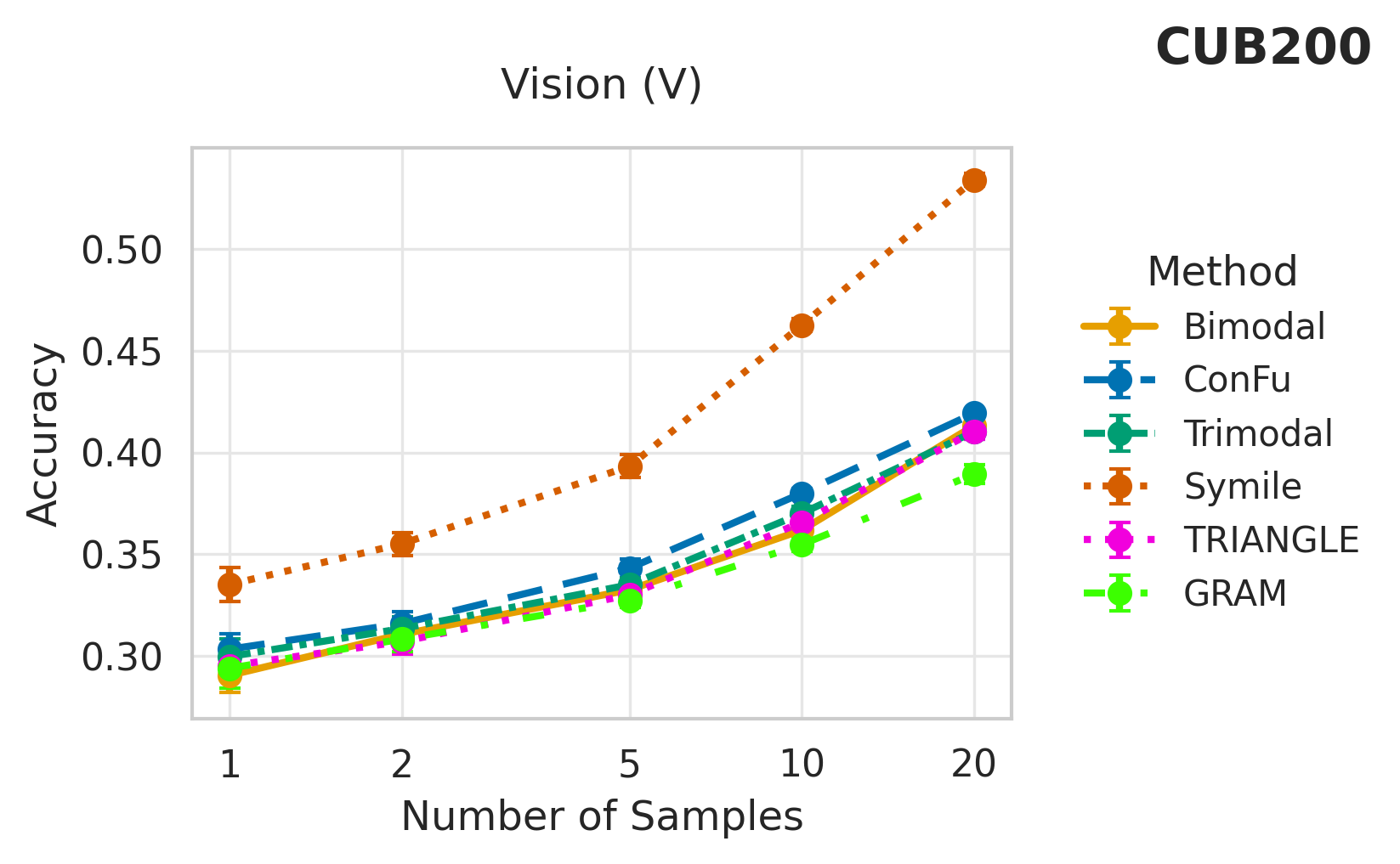}
    \caption{Few-shot linear probing results for CUB200~\cite{wah2011caltech}. Performance is shown as the number of labeled examples increases. Prediction is done in the single frame setting.}
    \label{fig:suppl_fewshot_cub200_singleframe}
\end{figure}

\renewcommand{\arraystretch}{1.1}
% \begin{table}[t]
% \caption{
% Zero-shot classification accuracy (\%) on SSW60 under different image augmentations.  
% V = Vision, AV = Audio-Visual. Best results in \textbf{bold}.
% }
% \centering
% \begin{tabular}{lcccc}
% \toprule
% \textbf{Method (Modality)} & \textbf{Low-res} & \textbf{Color} & \textbf{Blur} & \textbf{Noise} \\
% \midrule
% Tri-CLIP (V) & 43.31 & 62.04 & 47.37 & 35.43 \\
% \textbf{ConFu (AV)} & \textbf{51.43} & \textbf{66.74} & \textbf{54.78} & \textbf{45.17} \\
% \bottomrule
% \end{tabular}
% \label{tab:confu_vs_triclip_augmentations}
% \end{table}

% \clearpage

\section{Datasets}
\label{appendix:datasets}
\subsection{Bird-MML Dataset}
\label{appendix:datasets_bird_mml}
\begin{figure}[t]
    \centering
    \includegraphics[width=0.95\linewidth]{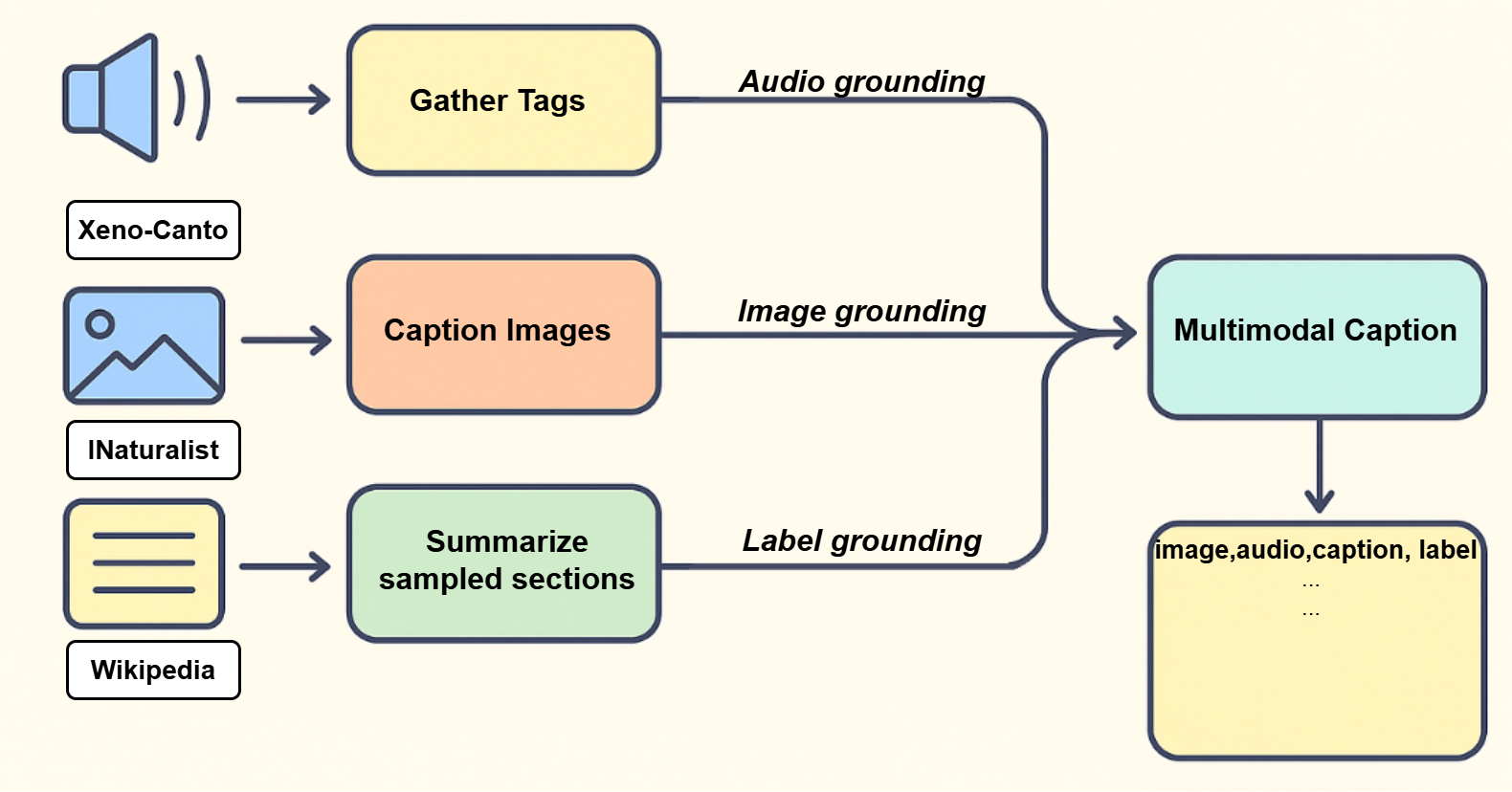}
    \caption{Overview of the multimodal data generation pipeline. Images and audio are collected from public sources (iNaturalist and Xeno-Canto), pseudocaptions are extracted from Wikipedia, and captions are generated via a combination of InstructBLIP2 and Gemma-2-2B-IT. The resulting triplets form the pretraining dataset.}
    \label{fig:data_pipeline}
\end{figure}

% \begin{figure*}[t]
%     \centering
%     \includegraphics[width=0.7\linewidth]{figs/supplementary_dataset/dataset_1.png} \\[4pt]
%     \includegraphics[width=0.7\linewidth]{figs/supplementary_dataset/dataset_2.png} \\[4pt]
%     \includegraphics[width=0.7\linewidth]{figs/supplementary_dataset/dataset_3.png}
%     \includegraphics[width=0.7\linewidth]{figs/supplementary_dataset/dataset_4.png} \\[4pt]
%     \includegraphics[width=0.7\linewidth]{figs/supplementary_dataset/dataset_5.png} \\[4pt]
%     % \includegraphics[width=0.8\linewidth]{figs/supplementary_dataset/dataset_6.png}
%     \caption{Examples from the constructed multimodal dataset, each triplet consisting of an image, its corresponding audio spectrogram, and a generated caption. The dataset integrates information from all three modalities, with captions often reflecting both visual and acoustic features as well as generic species-level information. While the image modality remains dominant in the descriptions, textual content also incorporates contextual cues inferred from the audio recordings.}
%     \label{fig:dataset_triplets}
% \end{figure*}

\noindent\textbf{Dataset Construction.}

To construct our dataset, we combined data from three sources: \textbf{iNaturalist} for images, \textbf{Xeno-Canto} for audio, and \textbf{Wikipedia} for textual and class-level grounding. For every bird class present in the VB100 or SSW60 datasets, we collected corresponding images (from iNaturalist), audio recordings (from Xeno-Canto), and Wikipedia articles.

\paragraph{Audio Processing}
Audio samples were segmented into 10-second clips (or zero-padded if shorter). We used the tags provided in Xeno-Canto as semantic grounding for the final captions.

\paragraph{Image Captioning}
Each image from iNaturalist was processed with \texttt{InstructBLIP2} to generate visual descriptions, which served as grounding for the final caption.

\noindent\textbf{Model:} \texttt{Salesforce/instructblip-flan-t5-xl}

\noindent\textbf{Prompt:}
\begin{quote}
\texttt{Describe the bird's colors, size, and shapes.}
\end{quote}

\paragraph{Wikipedia Text Processing}
For each class, we sampled random sections from the corresponding Wikipedia page. These sections were summarized by an LLM to produce textual grounding, generating 100 captions per class. This allowed us to capture diverse factual descriptions of each species.

\noindent\textbf{Model:} \texttt{google/gemma-2-2b-it}

\noindent\textbf{Prompt:}
\begin{quote}
\texttt{You are a naturalist assistant.}\\
\texttt{Summarize the following text into one coherent caption.}\\
\texttt{Be concise and factual.}\\[4pt]
\texttt{\{section\_text\}}\\[4pt]
\texttt{Caption:}
\end{quote}

\paragraph{Caption Combination \& Triplet Construction}
In the final step, we matched images, audio clips, and Wikipedia summaries into triplets. To produce a unified caption grounded in all three modalities (image, class, audio), we fed the image caption, audio tags, and a randomly selected Wikipedia caption into an LLM.

\noindent\textbf{Prompt:}
\begin{quote}
\texttt{You are a naturalist assistant.} \\
\texttt{Combine the following information into a concise caption (10--20 words).} \\
\texttt{Include information about the bird's appearance and sound if available.} \\
\texttt{Be factual and informative. \\[4pt]
Image caption: \{image info\}} \\
\texttt{Species caption: {\{textual info\}}} \\
\texttt{Sound tags: \{tags\}} \\[4pt]
\texttt{Caption:}
\end{quote}

This process results in coherent multimodal captions grounded jointly in visual appearance, species knowledge, and acoustic characteristics.

% \noindent\textbf{Dataset Samples.}
% Figure~\ref{fig:dataset_triplets} presents representative samples from our constructed multimodal dataset, where each triplet combines an image, its associated audio spectrogram, and a caption generated by a large language model (LLM). The captions often integrate cues from all three modalities, describing both the visual appearance of the bird and, in several cases, acoustic aspects such as calls or songs inferred from the audio tags associated with the respective audio sample. Despite this multimodal integration, the visual modality tends to dominate the descriptions, leading to image-centric narratives. Some captions also exhibit mild hallucinations or inconsistencies. Overall, the samples illustrate that the captions effectively merge visual and auditory context, albeit with varying degrees of grounding accuracy across modalities.

\subsection{Affective Computing Benchmarks}

MultiBench \cite{liang2021multibench} is a collection of benchmarking datasets designed to evaluate multimodal representation learning across a wide range of modalities and domains. In our experiments, we used the MOSI, UR-FUNNY, and MUStARD datasets. MultiBench provides pre-extracted features and predefined dataset splits, both of which we adopt in our setup.

MOSI is a multimodal sentiment analysis dataset consisting of 2,199 annotated YouTube video clips. UR-FUNNY contains 16,514 samples extracted from TED Talks and focuses on humor detection in spoken language. MUStARD includes 690 video clips from popular TV shows and is designed for multimodal sarcasm detection.

\subsection{AV-MNIST}
The AV-MNIST dataset is a multimodal benchmark that pairs degraded visual features with audio spectrograms to evaluate multimodal representation learning. The visual modality consists of 28×28 MNIST digits that have been PCA-projected, retaining only 25\% of the total variance. This dimensionality reduction is applied intentionally to weaken the visual signal and encourage effective fusion across modalities.

The audio modality comprises 112×112 spectrograms generated from the Free Spoken Digits Dataset~\cite{jackson2017fsdd}, with additive background noise sampled from ESC-50~\cite{piczak2015dataset} to increase variability and realism.

The dataset contains 55000 training samples and 10000 test samples. In our experiments, we use the pre-extracted audio spectrograms distributed by MultiBench \cite{liang2021multibench}.

% \clearpage

\section{Generalization to $M$ Modalities}
\label{app:generalM}
For the generalization of our framework to $M$ modalities, we extend the objective in Eq.~\ref{eq:final-loss} by summing InfoNCE losses over all relevant modality subsets. Specifically, we include all disjoint subset pairs
\begin{equation}
\mathcal{L}_{M}
=
\sum_{\substack{S_i,S_j \subseteq \{1,\dots,M\} \\
S_i \cap S_j = \emptyset,~S_i,S_j \neq \emptyset}}
\widehat{\mathcal{L}}_{\mathrm{InfoNCE}}^{(S_i,S_j)},
\label{eq:all-subsets}
\end{equation}
which correspond to mutual information terms $I(X_{S_i}; X_{S_j})$ estimated via InfoNCE bounds.  
This formulation provides a direct extension of our tri-modal framework, allowing contrastive learning to capture both pairwise and higher-order dependencies across arbitrary modality combinations.

To ascertain the above claim, we conducted preliminary experiments in a four-modality setting. This was achieved by splitting the original images into full spatial resolution monochromatic (grayscale) and low-resolution color (RGB), yielding four modalities alongside audio and text without additional data sources.  Although artificial, this scenario emulates the behavior of numerous remote sensing platforms (e.g., Pléiades: 0.5 meter monocromatic and 2 meters RGB) and the associated problem of pansharpening.
Experiments on VB100, and SSW60 show that the composite objective converges and consistently improves zero-shot performance over a pairwise-only baseline (Table~\ref{tab:4_modal_zero_shot_results_compact}), providing evidence that the approach remains effective as additional modalities are incorporated (with VB100 results reflecting the known weakness of the audio signal).

\begin{table}[h]
\centering
\caption{
Zero-shot classification accuracy (\%) on SSW60 and VB100 datasets.  
G = Grayscale Vision, LR = Low-Res RGB Vision, A = Audio. 
}
\resizebox{\columnwidth}{!}{%
\begin{tabular}{lcc@{\hskip 1em}cc}
\toprule
 & \multicolumn{2}{c}{\textbf{SSW60 Acc. (\%)}} & \multicolumn{2}{c}{\textbf{VB100 Acc. (\%)}} \\
\cmidrule(lr){2-3} \cmidrule(lr){4-5}
\textbf{Modality} & \textbf{Pairwise CLIP} & \textbf{ConFu} & \textbf{Pairwise CLIP} & \textbf{ConFu} \\
\midrule

G (grayscale)     & 50.33 & 48.46 & 14.48 & 15.47 \\
LR (low-res RGB)  & 58.28 & 55.22 & 15.89 & 16.74 \\
A (audio)        & 26.67 & 26.02 &  3.04 &  1.77 \\

\addlinespace
G + LR             & – & 59.39 & – & 18.15 \\
G + A             & –  & 52.20 & – & 14.05 \\
LR + A             & –  & 57.63 & – & 15.32 \\

\addlinespace
G + LR + A         & –     & \textbf{61.96} & –     & \textbf{15.32} \\

\bottomrule
\end{tabular}
}
\label{tab:4_modal_zero_shot_results_compact}
\end{table}

Nevertheless, while the formulation in Eq.~\ref{eq:all-subsets} captures all possible cross-subset dependencies, its combinatorial nature results in a large number of contrastive terms as $M$ increases. In practice, task-specific relaxations of this general objective can substantially reduce computational complexity. For instance \ref{eq:all-subsets-singleton}, when the goal is to retrieve a single target modality, one may restrict the loss to align any subset of the remaining modalities with that target only, i.e., terms of the form $I(X_t; X_S)$. Such relaxations preserve the representational alignment relevant to the retrieval task while avoiding the exponential growth in the number of InfoNCE terms, yielding a more tractable yet principled multimodal contrastive objective.

\begin{equation}
\mathcal{L}_{\mathrm{retrieval}}
=
\sum_{\substack{
S_i,S_j \subseteq \{1,\dots,M\} \\
S_i \cap S_j = \emptyset,\; S_i,S_j \neq \emptyset \\
|S_i| = 1
}}
\widehat{\mathcal{L}}_{\mathrm{InfoNCE}}^{(S_i,S_j)}.
\label{eq:all-subsets-singleton}
\end{equation}

\end{document}